\documentclass{article}
\usepackage[utf8]{inputenc}
\usepackage{amsmath}
\usepackage{amssymb}
\usepackage{algpseudocode}
\usepackage{algorithm}
\usepackage{enumitem}
\usepackage{graphicx}
\usepackage{tikz}
\usepackage{xcolor}
\usepackage{listings}
\usepackage{array}
\usepackage{xurl}
\usepackage[hidelinks]{hyperref}
\usetikzlibrary{decorations.pathreplacing}

\title{Detailed Derivation of The Backpropagation Algorithm}
\author{Faris Abdullah Alasmary \\
faris.alasmary@gmail.com}
\date{$27^{th}$ of October 2026}
\makeatletter
\providecommand{\theHALG@line}{\thealgorithm.\arabic{ALG@line}}
\makeatother

\lstnewenvironment{code}[1][]{\endgraf\medskip\noindent\minipage{\linewidth}\lstset{#1}}{\endminipage\endgraf\medskip}

\definecolor{codekeyword}{RGB}{0,0,170}
\definecolor{codecomment}{RGB}{0,128,0}
\definecolor{codestring}{RGB}{163,21,21}
\definecolor{codenumber}{RGB}{120,120,120}
\definecolor{codebg}{RGB}{248,248,248}
\newcommand{\drawnetwork}{
	\foreach \l/\n in {0/3, 1/5, 2/4, 3/3} {
		\foreach \j in {1,...,\n} {
			\node[circle, draw, minimum size=0.75cm, fill=white] (L\l-\j) at (3.2*\l, {1.15*((\n+1)/2-\j)}) {};
		}
	}
	\foreach \j in {1,...,3} { \foreach \k in {1,...,5} { \draw[gray!45] (L0-\j) -- (L1-\k); } }
	\foreach \j in {1,...,5} { \foreach \k in {1,...,4} { \draw[gray!45] (L1-\j) -- (L2-\k); } }
	\foreach \j in {1,...,4} { \foreach \k in {1,...,3} { \draw[gray!45] (L2-\j) -- (L3-\k); } }
	\foreach \l/\n in {0/3, 1/5, 2/4, 3/3} {
		\foreach \j in {1,...,\n} {
			\node[circle, draw, minimum size=0.75cm, fill=white] at (L\l-\j) {};
		}
	}
	\node[above=0.35cm] at (L0-1 |- L1-1) {$A^{[0]} = x$};
	\node[above=0.35cm] at (L1-1) {$A^{[1]}$};
	\node[above=0.35cm] at (L2-1 |- L1-1) {$A^{[2]}$};
	\node[above=0.35cm] at (L3-1 |- L1-1) {$A^{[3]} = \hat{y}$};
	\foreach \k in {1,2,3} { \node[right=0.5cm] at (L3-\k) {$E_{\k}$}; }
	\draw[decorate, decoration={brace, mirror, amplitude=6pt}] (0.1,-3.3) -- (3.1,-3.3) node[midway, below=8pt] {$\underbrace{W^{[1]}}_\text{$5\times 3$}$};
	\draw[decorate, decoration={brace, mirror, amplitude=6pt}] (3.3,-3.3) -- (6.3,-3.3) node[midway, below=8pt] {$\underbrace{W^{[2]}}_\text{$4\times 5$}$};
	\draw[decorate, decoration={brace, mirror, amplitude=6pt}] (6.5,-3.3) -- (9.5,-3.3) node[midway, below=8pt] {$\underbrace{W^{[3]}}_\text{$3\times 4$}$};
}

\begin{document}

	\maketitle

	\section{Notation}

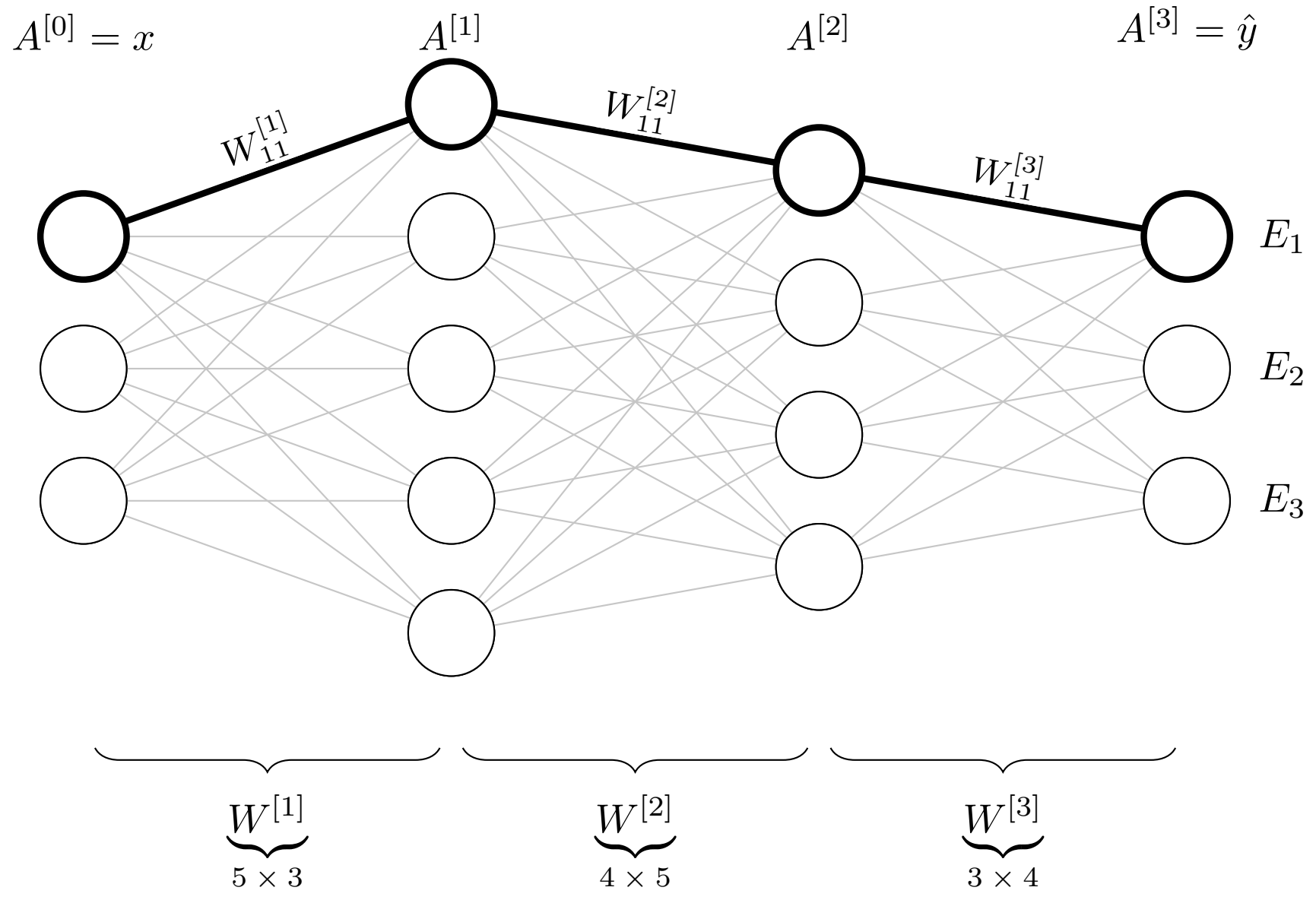
\begin{figure}[htbp]
		\centering
		\begin{tikzpicture}
			\drawnetwork
			\draw[line width=1.6pt] (L0-1) -- node[above, sloped, fill=white, inner sep=1pt] {\small $W^{[1]}_{11}$} (L1-1);
			\draw[line width=1.6pt] (L1-1) -- node[above, sloped, fill=white, inner sep=1pt] {\small $W^{[2]}_{11}$} (L2-1);
			\draw[line width=1.6pt] (L2-1) -- node[above, sloped, fill=white, inner sep=1pt] {\small $W^{[3]}_{11}$} (L3-1);
			\foreach \l in {0,1,2,3} { \node[circle, draw, line width=1.6pt, minimum size=0.75cm, fill=white] at (L\l-1) {}; }
		\end{tikzpicture}
		\caption{the running example network. The bold path connects the $1^{st}$ node of every layer to the error $E_{1}$ of the $1^{st}$ output node.}
		\label{network_figure}
	\end{figure}

	Here are some notation conventions:\\ \\
	$L =\:$ the number of layers in the network (the input layer is not counted).\\
	$n_{l} =\:$ the number of nodes (neurons) in the $l^{th}$ layer. The input layer is called layer $0$, so $n_{0}$ is the number of input values (features).\\
	$Z^{[l]} =\:$ a \textbf{vector} of the weighted sums of all the nodes in the $l^{th}$ layer (before applying the activation function).\\
	$Z^{[l]}_{j} =\:$ a \textbf{scalar} that represents the weighted sum of the $j^{th}$ node in the $l^{th}$ layer.\\
	$A^{[l]} =\:$ a \textbf{vector} of the outputs (activations) of all the nodes in the $l^{th}$ layer.\\
	$A^{[l]}_{j} =\:$ a \textbf{scalar} that represents the output of the $j^{th}$ node in the $l^{th}$ layer.\\
	$W^{[l]} =\:$ a \textbf{matrix} of the weights that connect the layer $(l-1)$ to the layer $l$.\\
	$W^{[l]}_{jk} =\:$ a \textbf{scalar} that represents the weight that connects the $k^{th}$ node of the layer $(l-1)$ to the $j^{th}$ node of the layer $l$.\\
	$b^{[l]} =\:$ a \textbf{vector} of the biases of the $l^{th}$ layer and $b^{[l]}_{j}$ is the bias of its $j^{th}$ node.\\
	$x^{(i)} =\:$ a \textbf{vector} of features of the $i^{th}$ example and $y^{(i)}$ is the vector of its target values.\\
	$\hat{y}^{(i)} =\:$ a \textbf{vector} of the predicted values of the $i^{th}$ example, i.e., the outputs of the last layer $A^{[L]}$.\\
	$m =\:$ the number of examples in the dataset.
	\\\\
	\textbf{Note}: the square brackets $[l]$ always refer to a \textbf{layer}, the parentheses $(i)$ always refer to an \textbf{example}, and the curly braces $\{t\}$ always refer to a \textbf{mini-batch} (Section(\ref{minibatch_section})). When we work with a single example, we will drop the superscript $(i)$ and simply write $x$, $y$, and $\hat{y}$.
	\\\\
	\textbf{Note}: for the subscripts, we will always use $j$ for a node of the current layer $l$, $k$ for a node of the previous layer $(l-1)$, and $p$ for a node of the next layer $(l+1)$.
	\\\\
	\textbf{Note}: the numbers under braces show the dimensions of matrices, i.e.,  $rows\times columns$.
	\\\\
	Each layer $l$ takes the outputs of the previous layer $A^{[l-1]}$ as its input. It multiplies them by its weights, adds its biases, and then passes the result to an activation function $\sigma$ (we will define it in Section(\ref{sigmoid_section})). The equation of a single layer is:
	\begin{equation}
	\begin{split}
	\underbrace{Z^{[l]}}_\text{$n_{l}\times 1$} &= \underbrace{W^{[l]}}_\text{$n_{l}\times n_{l-1}$} \cdot \underbrace{A^{[l-1]}}_\text{$n_{l-1}\times 1$} + \underbrace{b^{[l]}}_\text{$n_{l}\times 1$} \\
	\underbrace{A^{[l]}}_\text{$n_{l}\times 1$} &= \sigma\left(\underbrace{Z^{[l]}}_\text{$n_{l}\times 1$}\right)
	\end{split}
	\label{single_layer}
	\end{equation}
	If we look at a single node $j$ of the layer $l$, equation(\ref{single_layer}) becomes:
	\begin{equation}
	\begin{split}
	Z^{[l]}_{j} &= W^{[l]}_{j1}A^{[l-1]}_{1} + W^{[l]}_{j2}A^{[l-1]}_{2} + \dots + W^{[l]}_{jn_{l-1}}A^{[l-1]}_{n_{l-1}} + b^{[l]}_{j} \\
	&= \left(\sum_{k=1}^{n_{l-1}} W^{[l]}_{jk}A^{[l-1]}_{k}\right) + b^{[l]}_{j} \\
	A^{[l]}_{j} &= \sigma\left(Z^{[l]}_{j}\right)
	\end{split}
	\label{single_node}
	\end{equation}
	i.e., the node $j$ computes a weighted sum of all the outputs of the previous layer, exactly like the equation of a line in linear regression~\cite{alasmary_linear_regression}, and then applies the activation function to it. We will explain these equations in detail in Section(\ref{forward_section}).
	\\\\
	The weights matrix $W^{[l]}$ that is used in equation(\ref{single_layer}) is:
	\[
	W^{[l]} =
	\underbrace{
		\begin{bmatrix}
		W^{[l]}_{11}    & W^{[l]}_{12}    & W^{[l]}_{13}    & \dots  & W^{[l]}_{1n_{l-1}}    \\
		W^{[l]}_{21}    & W^{[l]}_{22}    & W^{[l]}_{23}    & \dots  & W^{[l]}_{2n_{l-1}}    \\
		W^{[l]}_{31}    & W^{[l]}_{32}    & W^{[l]}_{33}    & \dots  & W^{[l]}_{3n_{l-1}}    \\
		\vdots          & \vdots          & \vdots          & \ddots & \vdots                \\
		W^{[l]}_{n_{l}1} & W^{[l]}_{n_{l}2} & W^{[l]}_{n_{l}3} & \dots  & W^{[l]}_{n_{l}n_{l-1}}
		\end{bmatrix}}_\text{$n_{l}\times n_{l-1}$}
	\]
	In equation(\ref{single_node}), the weight $W^{[l]}_{jk}$ multiplies the output $A^{[l-1]}_{k}$ of the $k^{th}$ node of the layer $(l-1)$ and the result is added to $Z^{[l]}_{j}$ of the $j^{th}$ node of the layer $l$. Hence, the first index $j$ of $W^{[l]}_{jk}$ is the row number and it tells us \textbf{where the weight goes to} (the $j^{th}$ node of the layer $l$). The second index $k$ is the column number and it tells us \textbf{where the weight comes from} (the $k^{th}$ node of the layer $(l-1)$). This is the reason why the matrix $W^{[l]}$ has $n_{l}$ rows and $n_{l-1}$ columns. For example, $W^{[3]}_{21}$ is the weight that connects the $1^{st}$ node of the layer $2$ to the $2^{nd}$ node of the layer $3$.
	\\\\
	All the vectors in this paper are \textbf{column-vectors}. For example, the outputs of the layer $(l-1)$ are:
	\[
	A^{[l-1]} =
	\underbrace{
		\begin{bmatrix}
		A^{[l-1]}_{1} \\
		A^{[l-1]}_{2} \\
		A^{[l-1]}_{3} \\
		\vdots        \\
		A^{[l-1]}_{n_{l-1}}
		\end{bmatrix}}_\text{$n_{l-1}\times 1$}
	\]
	Throughout this paper, we will use the network shown in Figure(\ref{network_figure}) as a running example. It has $3$ input values, two hidden layers with $5$ and $4$ nodes, respectively, and an output layer with $3$ nodes, i.e., $L = 3$, $n_{0} = 3$, $n_{1} = 5$, $n_{2} = 4$, and $n_{3} = 3$. Each output node $j$ has its own error $E_{j}$ that we will define in Section(\ref{loss_section}).

	\section{Background}

	\subsection{The Sigmoid Function and Its Derivative}
	\label{sigmoid_section}
	Every node in our network computes a weighted sum $Z$ of its inputs and then passes it to an \textbf{activation function}. We will use the \textbf{sigmoid function} which is defined as:
	\begin{equation}
	\sigma(z) = \frac{1}{1 + e^{-z}}
	\label{sigmoid}
	\end{equation}
	The backpropagation algorithm needs the derivative of the activation function. We can write the sigmoid function as $\sigma(z) = \left(1 + e^{-z}\right)^{-1}$ and by applying the chain rule, we will get:
	\[
	\sigma'(z) = \frac{d\sigma(z)}{dz} = -1\cdot\left(1 + e^{-z}\right)^{-2}\cdot\left(-e^{-z}\right) = \frac{e^{-z}}{\left(1 + e^{-z}\right)^{2}}
	\]
	and by splitting the fraction into the product of two fractions
	\[
	\sigma'(z) = \frac{1}{1 + e^{-z}} \cdot \frac{e^{-z}}{1 + e^{-z}}
	\]
	The first fraction is $\sigma(z)$ itself. For the second fraction, we can add and subtract $1$ in the numerator:
	\[
	\frac{e^{-z}}{1 + e^{-z}} = \frac{\left(1 + e^{-z}\right) - 1}{1 + e^{-z}} = \frac{1 + e^{-z}}{1 + e^{-z}} - \frac{1}{1 + e^{-z}} = 1 - \sigma(z)
	\]
	hence,
	\begin{equation}
	\sigma'(z) = \sigma(z)\cdot\left(1 - \sigma(z)\right)
	\label{sigmoid_derivative}
	\end{equation}
	\textbf{Note}: when $\sigma$ or $\sigma'$ is applied to a vector, it is applied to each element of the vector separately (\textbf{element-wise}). For example,
	\[
	\sigma\left(
	\underbrace{
		\begin{bmatrix}
		Z_{1} \\
		Z_{2} \\
		Z_{3}
		\end{bmatrix}}_\text{$3\times 1$}
	\right)
	=
	\underbrace{
		\begin{bmatrix}
		\sigma(Z_{1}) \\
		\sigma(Z_{2}) \\
		\sigma(Z_{3})
		\end{bmatrix}}_\text{$3\times 1$},\;\;\;\;\;
	\sigma'\left(
	\underbrace{
		\begin{bmatrix}
		Z_{1} \\
		Z_{2} \\
		Z_{3}
		\end{bmatrix}}_\text{$3\times 1$}
	\right)
	=
	\underbrace{
		\begin{bmatrix}
		\sigma(Z_{1})\cdot\left(1 - \sigma(Z_{1})\right) \\
		\sigma(Z_{2})\cdot\left(1 - \sigma(Z_{2})\right) \\
		\sigma(Z_{3})\cdot\left(1 - \sigma(Z_{3})\right)
		\end{bmatrix}}_\text{$3\times 1$}
	\]

	\subsection{The Hadamard Product}
	The \textbf{Hadamard product}, denoted with the symbol $\odot$, is the element-wise product of two matrices (or vectors) that have the \textbf{same dimensions}. Each element of the result is the product of the two elements in the same position. For example,
	\[
	\underbrace{
		\begin{bmatrix}
		1 \\
		2 \\
		3 \\
		\end{bmatrix}}_\text{$3\times 1$}
	\;\odot\;
	\underbrace{
		\begin{bmatrix}
		4 \\
		5 \\
		6 \\
		\end{bmatrix}}_\text{$3\times 1$}
	\; = \;
	\underbrace{
		\begin{bmatrix}
		1\cdot 4 \\
		2\cdot 5 \\
		3\cdot 6 \\
		\end{bmatrix}}_\text{$3\times 1$}
	\; = \;
	\underbrace{
		\begin{bmatrix}
		4  \\
		10 \\
		18 \\
		\end{bmatrix}}_\text{$3\times 1$}
	\]
	\textbf{Note}: unlike the matrix product $\cdot$, the Hadamard product does not change the dimensions.

	\subsection{The Outer Product}
	If we multiply a column-vector $u$ by the transpose of another column-vector $v$, we will get a matrix. This product is called the \textbf{outer product}. For example, if we have the vectors
	\[
	u =
	\underbrace{
		\begin{bmatrix}
		u_{1} \\
		u_{2} \\
		u_{3}
		\end{bmatrix}}_\text{$3\times 1$},\;\;\;\;\;
	v =
	\underbrace{
		\begin{bmatrix}
		v_{1} \\
		v_{2}
		\end{bmatrix}}_\text{$2\times 1$}
	\]
	the outer product will be:
	\begin{equation}
	\underbrace{u}_\text{$3\times 1$} \cdot \underbrace{v^{T}}_\text{$1\times 2$}
	\; = \;
	\underbrace{
		\begin{bmatrix}
		u_{1} \\
		u_{2} \\
		u_{3}
		\end{bmatrix}}_\text{$3\times 1$}
	\cdot
	\underbrace{
		\begin{bmatrix}
		v_{1} & v_{2}
		\end{bmatrix}}_\text{$1\times 2$}
	\; = \;
	\underbrace{
		\begin{bmatrix}
		u_{1}v_{1} & u_{1}v_{2} \\
		u_{2}v_{1} & u_{2}v_{2} \\
		u_{3}v_{1} & u_{3}v_{2}
		\end{bmatrix}}_\text{$3\times 2$}
	\label{outer_product}
	\end{equation}
	We can see that the element in the $j^{th}$ row and the $k^{th}$ column of the result is $u_{j}v_{k}$.

	\subsection{Forward Propagation}
	\label{forward_section}
	The input layer does not compute anything. Its outputs are the features of the example:
	\[
	A^{[0]} = x
	\]
	Each node $j$ in the layer $l$ computes the weighted sum of the outputs of the previous layer plus its bias:
	\begin{equation}
	\begin{split}
	Z^{[l]}_{j} &= W^{[l]}_{j1}A^{[l-1]}_{1} + W^{[l]}_{j2}A^{[l-1]}_{2} + \dots + W^{[l]}_{jn_{l-1}}A^{[l-1]}_{n_{l-1}} + b^{[l]}_{j} \\
	&= \left(\sum_{k=1}^{n_{l-1}} W^{[l]}_{jk}A^{[l-1]}_{k}\right) + b^{[l]}_{j}
	\end{split}
	\label{weighted_sum_scalar}
	\end{equation}
	and then passes it to the activation function:
	\begin{equation}
	A^{[l]}_{j} = \sigma\left(Z^{[l]}_{j}\right)
	\label{activation_scalar}
	\end{equation}
	Equation(\ref{weighted_sum_scalar}) is very similar to the equation of a line that we used in linear regression. The difference is that each layer has many nodes and each node has its own weights. If we put all the weighted sums of the first layer of our network together, we can write them in the matrix form as follows:
			\[
	\underbrace{Z^{[1]}}_\text{$5\times 1$} = \underbrace{\underbrace{W^{[1]}}_\text{$5\times 3$} \cdot \underbrace{A^{[0]}}_\text{$3\times 1$}}_\text{$5\times 1$} + \underbrace{b^{[1]}}_\text{$5\times 1$}
	\]
	\[
	\underbrace{
		\begin{bmatrix}
		Z^{[1]}_{1} \\
		Z^{[1]}_{2} \\
		Z^{[1]}_{3} \\
		Z^{[1]}_{4} \\
		Z^{[1]}_{5}
		\end{bmatrix}}_\text{$5\times 1$}
	=
	\underbrace{
		\begin{bmatrix}
		W^{[1]}_{11} & W^{[1]}_{12} & W^{[1]}_{13} \\
		W^{[1]}_{21} & W^{[1]}_{22} & W^{[1]}_{23} \\
		W^{[1]}_{31} & W^{[1]}_{32} & W^{[1]}_{33} \\
		W^{[1]}_{41} & W^{[1]}_{42} & W^{[1]}_{43} \\
		W^{[1]}_{51} & W^{[1]}_{52} & W^{[1]}_{53}
		\end{bmatrix}}_\text{$5\times 3$}
	\cdot
	\underbrace{
		\begin{bmatrix}
		A^{[0]}_{1} \\
		A^{[0]}_{2} \\
		A^{[0]}_{3}
		\end{bmatrix}}_\text{$3\times 1$}
	+
	\underbrace{
		\begin{bmatrix}
		b^{[1]}_{1} \\
		b^{[1]}_{2} \\
		b^{[1]}_{3} \\
		b^{[1]}_{4} \\
		b^{[1]}_{5}
		\end{bmatrix}}_\text{$5\times 1$}
	\]
	\[
	\underbrace{
		\begin{bmatrix}
		Z^{[1]}_{1} \\
		Z^{[1]}_{2} \\
		Z^{[1]}_{3} \\
		Z^{[1]}_{4} \\
		Z^{[1]}_{5}
		\end{bmatrix}}_\text{$5\times 1$}
	=
	\underbrace{
		\begin{bmatrix}
		W^{[1]}_{11}A^{[0]}_{1} + W^{[1]}_{12}A^{[0]}_{2} + W^{[1]}_{13}A^{[0]}_{3} + b^{[1]}_{1} \\
		W^{[1]}_{21}A^{[0]}_{1} + W^{[1]}_{22}A^{[0]}_{2} + W^{[1]}_{23}A^{[0]}_{3} + b^{[1]}_{2} \\
		W^{[1]}_{31}A^{[0]}_{1} + W^{[1]}_{32}A^{[0]}_{2} + W^{[1]}_{33}A^{[0]}_{3} + b^{[1]}_{3} \\
		W^{[1]}_{41}A^{[0]}_{1} + W^{[1]}_{42}A^{[0]}_{2} + W^{[1]}_{43}A^{[0]}_{3} + b^{[1]}_{4} \\
		W^{[1]}_{51}A^{[0]}_{1} + W^{[1]}_{52}A^{[0]}_{2} + W^{[1]}_{53}A^{[0]}_{3} + b^{[1]}_{5}
		\end{bmatrix}}_\text{$5\times 1$}
	=
	\underbrace{
		\begin{bmatrix}
		\sum_{k=1}^{3} W^{[1]}_{1k}A^{[0]}_{k} + b^{[1]}_{1} \\
		\sum_{k=1}^{3} W^{[1]}_{2k}A^{[0]}_{k} + b^{[1]}_{2} \\
		\sum_{k=1}^{3} W^{[1]}_{3k}A^{[0]}_{k} + b^{[1]}_{3} \\
		\sum_{k=1}^{3} W^{[1]}_{4k}A^{[0]}_{k} + b^{[1]}_{4} \\
		\sum_{k=1}^{3} W^{[1]}_{5k}A^{[0]}_{k} + b^{[1]}_{5}
		\end{bmatrix}}_\text{$5\times 1$}
	\]
    and the outputs of the first layer are:
	\[
	\underbrace{A^{[1]}}_\text{$5\times 1$} = \sigma\left(\underbrace{Z^{[1]}}_\text{$5\times 1$}\right) =
	\underbrace{
		\begin{bmatrix}
		\sigma(Z^{[1]}_{1}) \\
		\sigma(Z^{[1]}_{2}) \\
		\sigma(Z^{[1]}_{3}) \\
		\sigma(Z^{[1]}_{4}) \\
		\sigma(Z^{[1]}_{5})
		\end{bmatrix}}_\text{$5\times 1$}
	\]
	The same thing is done for the second and the third layers:
	\[
	\begin{split}
	\underbrace{Z^{[2]}}_\text{$4\times 1$} &= \underbrace{\underbrace{W^{[2]}}_\text{$4\times 5$} \cdot \underbrace{A^{[1]}}_\text{$5\times 1$}}_\text{$4\times 1$} + \underbrace{b^{[2]}}_\text{$4\times 1$} \\
	\underbrace{A^{[2]}}_\text{$4\times 1$} &= \sigma\left(\underbrace{Z^{[2]}}_\text{$4\times 1$}\right)
	\end{split}
	\]
	\[
	\begin{split}
	\underbrace{Z^{[3]}}_\text{$3\times 1$} &= \underbrace{\underbrace{W^{[3]}}_\text{$3\times 4$} \cdot \underbrace{A^{[2]}}_\text{$4\times 1$}}_\text{$3\times 1$} + \underbrace{b^{[3]}}_\text{$3\times 1$} \\
	\underbrace{A^{[3]}}_\text{$3\times 1$} &= \sigma\left(\underbrace{Z^{[3]}}_\text{$3\times 1$}\right)
	\end{split}
	\]
	In general, the \textbf{forward propagation} of any layer $l$ is:
	\begin{equation}
	\boxed{
	\begin{split}
	\underbrace{A^{[0]}}_\text{$n_{0}\times 1$} &= \underbrace{x}_\text{$n_{0}\times 1$} \\
	\underbrace{Z^{[l]}}_\text{$n_{l}\times 1$} &= \underbrace{W^{[l]}}_\text{$n_{l}\times n_{l-1}$} \cdot \underbrace{A^{[l-1]}}_\text{$n_{l-1}\times 1$} + \underbrace{b^{[l]}}_\text{$n_{l}\times 1$} \\
	\underbrace{A^{[l]}}_\text{$n_{l}\times 1$} &= \sigma\left(\underbrace{Z^{[l]}}_\text{$n_{l}\times 1$}\right)
	\end{split}
	}
	\label{forward_propagation}
	\end{equation}
	\subsection{The Loss and The Cost Functions}
	\label{loss_section}
	The outputs of the last layer are the predicted values of the network. Therefore, we will write them as $\hat{y}$:
	\[
	\underbrace{\hat{y}}_\text{$3\times 1$} = \underbrace{A^{[3]}}_\text{$3\times 1$},\;\;\;\;\;
	\hat{y}_{j} = A^{[3]}_{j}
	\]
	To measure how close the prediction $\hat{y}_{j}$ of the $j^{th}$ output node is to its target value $y_{j}$, we will use the squared residual error that we used in linear regression. The error of the $j^{th}$ output node is:
	\begin{equation}
	E_{j} = Loss\left(y_{j}, \hat{y}_{j}\right) = \frac{1}{2}\cdot\left(y_{j} - \hat{y}_{j}\right)^{2}
	\label{loss_function}
	\end{equation}
	\textbf{Note}: the $\frac{1}{2}$ is added only to cancel the $2$ that comes from the derivative of the square. It does not change the location of the minimum.
	\\\\
	The total error of the network for a single example is the sum of the errors of all output nodes:
	\begin{equation}
	E = E_{1} + E_{2} + E_{3} = \frac{1}{2}\sum_{j=1}^{3}\left(y_{j} - \hat{y}_{j}\right)^{2} =
	\frac{1}{2}\cdot
	\underbrace{\underbrace{
		\left(y - \hat{y}\right)^{T}
	}_\text{$1\times 3$}
	\cdot
	\underbrace{
		\left(y - \hat{y}\right)
	}_\text{$3\times 1$}}_\text{$1\times 1$}
	\label{total_error}
	\end{equation}
	which is the same $e^{T}\cdot e$ form of the residual sum of squares that we derived in linear regression, where $e = y - \hat{y}$ is the vector of residual errors. Finally, if we have $m$ examples, the \textbf{cost function} $J$ is the average of the errors of all examples, where $E^{(i)}$ is the total error $E$ of the $i^{th}$ example:
	\begin{equation}
	J = \frac{1}{m}\sum_{i=1}^{m} E^{(i)} = \frac{1}{m}\sum_{i=1}^{m} \frac{1}{2}\cdot\left(y^{(i)} - \hat{y}^{(i)}\right)^{T}\cdot\left(y^{(i)} - \hat{y}^{(i)}\right)
	\label{cost_function}
	\end{equation}
    \newpage
	\subsection{Why Not a Closed-Form Solution?}
	\label{closed_form_section}
	\textbf{Our Goal}: we want to find the weights $W^{[1]}, W^{[2]}, W^{[3]}$ and the biases $b^{[1]}, b^{[2]}, b^{[3]}$ such that the cost $J$ is as small as possible.
	\\\\
	In linear regression~\cite{alasmary_linear_regression}, we found the best weights directly with the normal equation (closed-form solution). We could do that because the prediction was a \textbf{linear} function of the weights:
	\[
	\underbrace{\hat{y}}_\text{$m\times 1$} = \underbrace{X}_\text{$m\times (n+1)$} \cdot \underbrace{w}_\text{$(n+1)\times 1$}
	\]
	\textbf{Note}: here $X$, $y$, and $n$ have their meanings from the linear regression paper~\cite{alasmary_linear_regression}: $X$ has the examples as \textbf{rows}, $y$ holds the targets of all $m$ examples, and $n$ is the number of features. In this paper, $X$ will have the examples as \textbf{columns} (Section(\ref{batch_section})), $y$ is the target vector of a single example, and the number of features is $n_{0}$.
	\\\\
	Hence, the $RSS$ was a quadratic function of $w$ and its gradient was a linear function of $w$. When we set the gradient to zero, we got a system of linear equations that we solved with the inverse matrix:
	\[
	\nabla_{w}{RSS} = 2\cdot X^{T} \cdot X \cdot w - 2\cdot X^{T} \cdot y = 0 \;\;\Longrightarrow\;\; w = (X^{T} \cdot X)^{-1} \cdot X^{T} \cdot y
	\]
	In our network, if we substitute each layer of the forward propagation equation(\ref{forward_propagation}) into the next one, the prediction becomes:
	\begin{equation}
	\hat{y} = A^{[3]} = \sigma\left(W^{[3]} \cdot \sigma\left(W^{[2]} \cdot \sigma\left(W^{[1]} \cdot x + b^{[1]}\right) + b^{[2]}\right) + b^{[3]}\right)
	\label{nested_prediction}
	\end{equation}
	We can see clearly that $\hat{y}$ is \textbf{not} a linear function of the weights. The weights are inside sigmoid functions, i.e., inside exponentials $e^{-z}$, and these sigmoid functions are nested inside each other. This causes two problems:
	\begin{enumerate}[leftmargin=1cm, label=\arabic*.]
		\item If we set the gradient to zero, we will get a system of \textbf{nonlinear} equations. For example, we will see in Section(\ref{derivation_section}) that one of these equations is:
		\[
		\frac{\partial E}{\partial W^{[3]}_{11}} = -\left(y_{1} - \sigma\left(Z^{[3]}_{1}\right)\right)\cdot\sigma\left(Z^{[3]}_{1}\right)\cdot\left(1 - \sigma\left(Z^{[3]}_{1}\right)\right)\cdot A^{[2]}_{1} = 0
		\]
		where $Z^{[3]}_{1}$ and $A^{[2]}_{1}$ depend on all the weights of the previous layers. Unlike $X^{T} \cdot X \cdot w = X^{T} \cdot y$, we cannot move the weights to one side and take an inverse matrix, since the weights cannot be separated from the exponentials. There is no algebraic formula that solves these equations.
		\item Even if we could solve them, the solution would not be unique. The cost $J$ of linear regression is shaped like a bowl, so it has only one point where the gradient is zero, which is the minimum. The cost $J$ of a neural network is \textbf{not} shaped like a bowl. It has many points where the gradient is zero: many local minima and saddle points. For example, if we swap two nodes of a hidden layer together with their weights, the network will give exactly the same outputs and the same cost, so every minimum has many copies.
	\end{enumerate}
	Therefore, we will use gradient descent instead. Gradient descent does not need to solve $\nabla J = 0$. As we saw in linear regression, it only needs to \textbf{evaluate} the gradient at the current weights and then take a small step in the opposite direction. This means that we need the partial derivative of the error with respect to \textbf{every single weight and bias} in the network. The \textbf{backpropagation algorithm} is an efficient way to compute all these partial derivatives.
    
	\section{Mathematical Derivation}
    \label{derivation_section}
	We will first derive the partial derivatives for a single example and then we will extend them to $m$ examples. We will start from the output layer and move backward towards the input layer, hence the name \textbf{back}propagation.
	\\\\
	\textbf{Note}: to keep track of the dimensions of the matrices, we will show the dimensions under each term.

	\subsection{The Output Layer (Layer 3)}
	Let's start with the weight $W^{[3]}_{11}$ which is the last weight on the bold path in Figure(\ref{network_figure}). This weight affects the error $E_{1}$ through the following chain:
	\[
	W^{[3]}_{11} \;\longrightarrow\; Z^{[3]}_{1} \;\longrightarrow\; \hat{y}_{1} \;\longrightarrow\; E_{1}
	\]
	The weight $W^{[3]}_{11}$ appears only in $Z^{[3]}_{1}$. It does not appear in $Z^{[3]}_{2}$ or $Z^{[3]}_{3}$, so it does not affect $E_{2}$ or $E_{3}$. Hence, $\frac{\partial E}{\partial W^{[3]}_{11}} = \frac{\partial E_{1}}{\partial W^{[3]}_{11}}$ and by applying the chain rule, we will get:
	\begin{equation}
	\frac{\partial E_{1}}{\partial W^{[3]}_{11}} =
	\frac{\partial E_{1}}{\partial \hat{y}_{1}} \cdot
	\frac{\partial \hat{y}_{1}}{\partial Z^{[3]}_{1}} \cdot
	\frac{\partial Z^{[3]}_{1}}{\partial W^{[3]}_{11}}
	\label{chain_rule_layer3}
	\end{equation}
	We will compute each of the three terms separately. The first term is the derivative of the loss function in equation(\ref{loss_function}):
	\begin{equation}
	\frac{\partial E_{1}}{\partial \hat{y}_{1}} = \frac{\partial}{\partial \hat{y}_{1}}\left(\frac{1}{2}\cdot\left(y_{1} - \hat{y}_{1}\right)^{2}\right) = 2\cdot\frac{1}{2}\cdot\left(y_{1} - \hat{y}_{1}\right)\cdot\left(-1\right) = -\left(y_{1} - \hat{y}_{1}\right)
	\label{loss_derivative}
	\end{equation}
	The second term is the derivative of the sigmoid function in equation(\ref{sigmoid_derivative}), since $\hat{y}_{1} = A^{[3]}_{1} = \sigma\left(Z^{[3]}_{1}\right)$:
	\begin{equation}
	\frac{\partial \hat{y}_{1}}{\partial Z^{[3]}_{1}} = \sigma'\left(Z^{[3]}_{1}\right) = \sigma\left(Z^{[3]}_{1}\right)\cdot\left(1 - \sigma\left(Z^{[3]}_{1}\right)\right)
	\label{activation_derivative_layer3}
	\end{equation}
	For the third term, we know from equation(\ref{weighted_sum_scalar}) that
	\[
	Z^{[3]}_{1} = \sum_{k=1}^{4} W^{[3]}_{1k}A^{[2]}_{k} + b^{[3]}_{1} = W^{[3]}_{11}A^{[2]}_{1} + W^{[3]}_{12}A^{[2]}_{2} + W^{[3]}_{13}A^{[2]}_{3} + W^{[3]}_{14}A^{[2]}_{4} + b^{[3]}_{1}
	\]
	and all the terms are constants with respect to $W^{[3]}_{11}$ except the first one, hence,
	\begin{equation}
	\frac{\partial Z^{[3]}_{1}}{\partial W^{[3]}_{11}} = \frac{\partial}{\partial W^{[3]}_{11}}\left(\sum_{k=1}^{4} W^{[3]}_{1k}A^{[2]}_{k} + b^{[3]}_{1}\right) = A^{[2]}_{1}
	\label{z_derivative_layer3}
	\end{equation}
	The first two terms of equation(\ref{chain_rule_layer3}) depend only on the output node and not on the weight. We will give their product a special name, $\delta^{[3]}_{1}$, and we will call it the \textbf{error} of the $1^{st}$ node of the layer $3$:
	\begin{equation}
	\boxed{
	\delta^{[3]}_{1} = \frac{\partial E_{1}}{\partial \hat{y}_{1}} \cdot \frac{\partial \hat{y}_{1}}{\partial Z^{[3]}_{1}} = -\left(y_{1} - \hat{y}_{1}\right)\cdot\sigma'\left(Z^{[3]}_{1}\right)
	}
	\label{delta_3_1}
	\end{equation}
	\textbf{Note}: by the chain rule, $\delta^{[3]}_{1}$ is simply $\frac{\partial E}{\partial Z^{[3]}_{1}}$, i.e., how much the total error changes when the weighted sum $Z^{[3]}_{1}$ changes.
	\\\\
	From equation(\ref{delta_3_1}) and equation(\ref{z_derivative_layer3}), we can rewrite equation(\ref{chain_rule_layer3}) as follows:
	\begin{equation}
	\begin{split}
	\frac{\partial E}{\partial W^{[3]}_{11}} &= \delta^{[3]}_{1} \cdot \frac{\partial Z^{[3]}_{1}}{\partial W^{[3]}_{11}} = \delta^{[3]}_{1} \cdot A^{[2]}_{1} \\
	&= -\left(y_{1} - \hat{y}_{1}\right)\cdot\sigma\left(Z^{[3]}_{1}\right)\cdot\left(1 - \sigma\left(Z^{[3]}_{1}\right)\right)\cdot A^{[2]}_{1}
	\end{split}
	\label{dE_dW3_11}
	\end{equation}
	If we repeat the same steps for any weight $W^{[3]}_{jk}$ that connects the $k^{th}$ node of the layer $2$ to the $j^{th}$ node of the layer $3$, we will get:
	\begin{equation}
	\delta^{[3]}_{j} = -\left(y_{j} - \hat{y}_{j}\right)\cdot\sigma'\left(Z^{[3]}_{j}\right),\;\;\;\;\;
	\frac{\partial E}{\partial W^{[3]}_{jk}} = \delta^{[3]}_{j} \cdot A^{[2]}_{k}
	\label{dE_dW3_jk}
	\end{equation}
	Now, let's put the errors of all the output nodes together in a single vector:
	\[
	\underbrace{\delta^{[3]}}_\text{$3\times 1$} =
	\underbrace{
		\begin{bmatrix}
		\delta^{[3]}_{1} \\
		\delta^{[3]}_{2} \\
		\delta^{[3]}_{3}
		\end{bmatrix}}_\text{$3\times 1$}
	=
	\underbrace{
		\begin{bmatrix}
		-\left(y_{1} - \hat{y}_{1}\right)\cdot\sigma'\left(Z^{[3]}_{1}\right) \\
		-\left(y_{2} - \hat{y}_{2}\right)\cdot\sigma'\left(Z^{[3]}_{2}\right) \\
		-\left(y_{3} - \hat{y}_{3}\right)\cdot\sigma'\left(Z^{[3]}_{3}\right)
		\end{bmatrix}}_\text{$3\times 1$}
	=
	\underbrace{
		\begin{bmatrix}
		-\left(y_{1} - \hat{y}_{1}\right) \\
		-\left(y_{2} - \hat{y}_{2}\right) \\
		-\left(y_{3} - \hat{y}_{3}\right)
		\end{bmatrix}}_\text{$3\times 1$}
	\odot
	\underbrace{
		\begin{bmatrix}
		\sigma'\left(Z^{[3]}_{1}\right) \\
		\sigma'\left(Z^{[3]}_{2}\right) \\
		\sigma'\left(Z^{[3]}_{3}\right)
		\end{bmatrix}}_\text{$3\times 1$}
	\]
	hence,
	\begin{equation}
	\underbrace{\delta^{[3]}}_\text{$3\times 1$} =
	\underbrace{-\left(y - \hat{y}\right)}_\text{$3\times 1$}
	\odot
	\underbrace{\sigma'\left(Z^{[3]}\right)}_\text{$3\times 1$}
	\label{delta_3_vector}
	\end{equation}
	Similarly, let's put all the partial derivatives of the weights of the layer $3$ in a matrix that has the same dimensions as $W^{[3]}$. By using equation(\ref{dE_dW3_jk}) for every element, we will get:
	\[
	\underbrace{\frac{\partial E}{\partial W^{[3]}}}_\text{$3\times 4$} =
	\underbrace{
		\begin{bmatrix}
		\frac{\partial E}{\partial W^{[3]}_{11}} & \frac{\partial E}{\partial W^{[3]}_{12}} & \frac{\partial E}{\partial W^{[3]}_{13}} & \frac{\partial E}{\partial W^{[3]}_{14}} \\[1ex]
		\frac{\partial E}{\partial W^{[3]}_{21}} & \frac{\partial E}{\partial W^{[3]}_{22}} & \frac{\partial E}{\partial W^{[3]}_{23}} & \frac{\partial E}{\partial W^{[3]}_{24}} \\[1ex]
		\frac{\partial E}{\partial W^{[3]}_{31}} & \frac{\partial E}{\partial W^{[3]}_{32}} & \frac{\partial E}{\partial W^{[3]}_{33}} & \frac{\partial E}{\partial W^{[3]}_{34}}
		\end{bmatrix}}_\text{$3\times 4$}
	\]
	\[
	=
	\underbrace{
		\begin{bmatrix}
		\delta^{[3]}_{1}A^{[2]}_{1} & \delta^{[3]}_{1}A^{[2]}_{2} & \delta^{[3]}_{1}A^{[2]}_{3} & \delta^{[3]}_{1}A^{[2]}_{4} \\
		\delta^{[3]}_{2}A^{[2]}_{1} & \delta^{[3]}_{2}A^{[2]}_{2} & \delta^{[3]}_{2}A^{[2]}_{3} & \delta^{[3]}_{2}A^{[2]}_{4} \\
		\delta^{[3]}_{3}A^{[2]}_{1} & \delta^{[3]}_{3}A^{[2]}_{2} & \delta^{[3]}_{3}A^{[2]}_{3} & \delta^{[3]}_{3}A^{[2]}_{4}
		\end{bmatrix}}_\text{$3\times 4$}
	\]
	The element in the $j^{th}$ row and the $k^{th}$ column is $\delta^{[3]}_{j}A^{[2]}_{k}$ which is exactly the outer product in equation(\ref{outer_product}):
	\[
	=
	\underbrace{\underbrace{
		\begin{bmatrix}
		\delta^{[3]}_{1} \\
		\delta^{[3]}_{2} \\
		\delta^{[3]}_{3}
		\end{bmatrix}}_\text{$3\times 1$}
	\cdot
	\underbrace{
		\begin{bmatrix}
		A^{[2]}_{1} & A^{[2]}_{2} & A^{[2]}_{3} & A^{[2]}_{4}
		\end{bmatrix}}_\text{$1\times 4$}}_\text{$3\times 4$}
	\]
	hence,
	\begin{equation}
	\underbrace{\frac{\partial E}{\partial W^{[3]}}}_\text{$3\times 4$} = \underbrace{\underbrace{\delta^{[3]}}_\text{$3\times 1$} \cdot \underbrace{A^{[2]T}}_\text{$1\times 4$}}_\text{$3\times 4$}
	\label{dE_dW3}
	\end{equation}
	For the biases, we can see from equation(\ref{weighted_sum_scalar}) that $\frac{\partial Z^{[3]}_{j}}{\partial b^{[3]}_{j}} = 1$. If we replace the third term of the chain rule in equation(\ref{chain_rule_layer3}) with this value, we will get $\frac{\partial E}{\partial b^{[3]}_{j}} = \delta^{[3]}_{j}\cdot 1 = \delta^{[3]}_{j}$, i.e.,
	\begin{equation}
	\underbrace{\frac{\partial E}{\partial b^{[3]}}}_\text{$3\times 1$} = \underbrace{\delta^{[3]}}_\text{$3\times 1$}
	\label{dE_db3}
	\end{equation}

	\newpage
	\subsection{The Second Hidden Layer (Layer 2)}
\begin{figure}[htbp]
		\centering
		\begin{tikzpicture}
			\drawnetwork
			\draw[line width=1.6pt] (L1-1) -- node[above, sloped, fill=white, inner sep=1pt] {\small $W^{[2]}_{11}$} (L2-1);
			\draw[line width=1.6pt] (L2-1) -- node[pos=0.55, above, sloped, inner sep=2pt] {\small $W^{[3]}_{11}$} (L3-1);
			\draw[line width=1.6pt] (L2-1) -- node[pos=0.72, above, sloped, inner sep=2pt] {\small $W^{[3]}_{21}$} (L3-2);
			\draw[line width=1.6pt] (L2-1) -- node[pos=0.62, below, sloped, inner sep=2pt] {\small $W^{[3]}_{31}$} (L3-3);
			\foreach \n in {L1-1, L2-1, L3-1, L3-2, L3-3} { \node[circle, draw, line width=1.6pt, minimum size=0.75cm, fill=white] at (\n) {}; }
		\end{tikzpicture}
		\caption{the three paths from the weight $W^{[2]}_{11}$ to the errors $E_{1}$, $E_{2}$, and $E_{3}$.}
		\label{paths_figure}
	\end{figure}
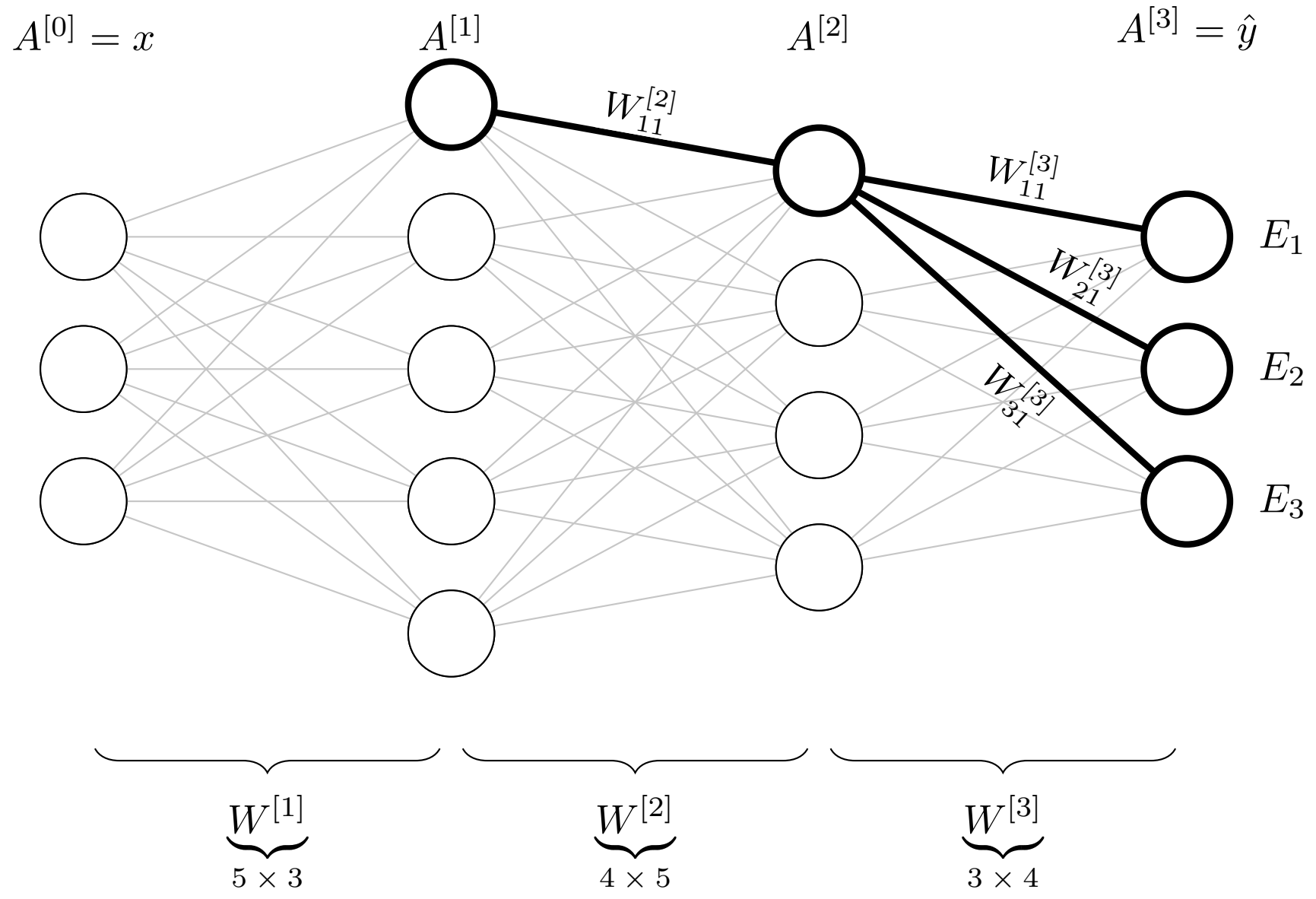

	Now, let's move one step backward to the weight $W^{[2]}_{11}$ in Figure(\ref{network_figure}). If we follow the bold path, this weight affects the error $E_{1}$ through the following chain:
	\[
	W^{[2]}_{11} \;\longrightarrow\; Z^{[2]}_{1} \;\longrightarrow\; A^{[2]}_{1} \;\longrightarrow\; Z^{[3]}_{1} \;\longrightarrow\; \hat{y}_{1} \;\longrightarrow\; E_{1}
	\]
	and by applying the chain rule, we will get:
	\begin{equation}
	\frac{\partial E_{1}}{\partial W^{[2]}_{11}} =
	\underbrace{
		\frac{\partial E_{1}}{\partial \hat{y}_{1}} \cdot
		\frac{\partial \hat{y}_{1}}{\partial Z^{[3]}_{1}}
	}_{\delta^{[3]}_{1}}
	\cdot
	\frac{\partial Z^{[3]}_{1}}{\partial A^{[2]}_{1}} \cdot
	\frac{\partial A^{[2]}_{1}}{\partial Z^{[2]}_{1}} \cdot
	\frac{\partial Z^{[2]}_{1}}{\partial W^{[2]}_{11}}
	\label{chain_rule_layer2}
	\end{equation}
	The first two terms are $\delta^{[3]}_{1}$ that we computed in equation(\ref{delta_3_1}). We do not need to compute them again. Hence,
	\[
	\frac{\partial E_{1}}{\partial W^{[2]}_{11}} = \delta^{[3]}_{1} \cdot
	\overbrace{
		\frac{\partial Z^{[3]}_{1}}{\partial A^{[2]}_{1}}
	}^{weight}
	\cdot
	\overbrace{
		\frac{\partial A^{[2]}_{1}}{\partial Z^{[2]}_{1}}
	}^{activation's\;derivative}
	\cdot
	\frac{\partial Z^{[2]}_{1}}{\partial W^{[2]}_{11}}
	\]
	The third term is the derivative of $Z^{[3]}_{1}$ with respect to $A^{[2]}_{1}$ (not with respect to the weight as before). In the sum $Z^{[3]}_{1} = \sum_{k=1}^{4} W^{[3]}_{1k}A^{[2]}_{k} + b^{[3]}_{1}$, only the term $W^{[3]}_{11}A^{[2]}_{1}$ contains $A^{[2]}_{1}$, hence,
	\begin{equation}
	\frac{\partial Z^{[3]}_{1}}{\partial A^{[2]}_{1}} = \frac{\partial}{\partial A^{[2]}_{1}}\left(\sum_{k=1}^{4} W^{[3]}_{1k}A^{[2]}_{k} + b^{[3]}_{1}\right) = W^{[3]}_{11}
	\label{z3_derivative_a2}
	\end{equation}
	The fourth term is the derivative of the sigmoid function:
	\begin{equation}
	\frac{\partial A^{[2]}_{1}}{\partial Z^{[2]}_{1}} = \sigma'\left(Z^{[2]}_{1}\right) = \sigma\left(Z^{[2]}_{1}\right)\cdot\left(1 - \sigma\left(Z^{[2]}_{1}\right)\right)
	\label{activation_derivative_layer2}
	\end{equation}
	and the last term can be computed in the same way as equation(\ref{z_derivative_layer3}):
	\begin{equation}
	\frac{\partial Z^{[2]}_{1}}{\partial W^{[2]}_{11}} = \frac{\partial}{\partial W^{[2]}_{11}}\left(\sum_{k=1}^{5} W^{[2]}_{1k}A^{[1]}_{k} + b^{[2]}_{1}\right) = A^{[1]}_{1}
	\label{z_derivative_layer2}
	\end{equation}
	By substituting equations(\ref{z3_derivative_a2}), (\ref{activation_derivative_layer2}), and (\ref{z_derivative_layer2}) into equation(\ref{chain_rule_layer2}), we will get:
	\begin{equation}
	\frac{\partial E_{1}}{\partial W^{[2]}_{11}} = \delta^{[3]}_{1} \cdot W^{[3]}_{11} \cdot \sigma'\left(Z^{[2]}_{1}\right) \cdot A^{[1]}_{1}
	\label{dE1_dW2_11}
	\end{equation}
	\textbf{But}, this is not the complete derivative of the total error $E$. In the output layer, the weight $W^{[3]}_{11}$ affected only $E_{1}$. Here, the weight $W^{[2]}_{11}$ changes $A^{[2]}_{1}$, and $A^{[2]}_{1}$ is an input to \textbf{every} node of the layer $3$ as shown in Figure(\ref{paths_figure}). Hence, $W^{[2]}_{11}$ affects $E_{1}$, $E_{2}$, and $E_{3}$ through three different paths.

	Since the total error is $E = E_{1} + E_{2} + E_{3}$ and the derivative of a sum is the sum of the derivatives:
	\[
	\frac{\partial E}{\partial W^{[2]}_{11}} = \frac{\partial E_{1}}{\partial W^{[2]}_{11}} + \frac{\partial E_{2}}{\partial W^{[2]}_{11}} + \frac{\partial E_{3}}{\partial W^{[2]}_{11}}
	\]
	The path to $E_{2}$ goes through the $2^{nd}$ output node instead of the $1^{st}$ one, so we can repeat the derivation of equation(\ref{dE1_dW2_11}) by replacing every index $1$ of the layer $3$ with $2$. The derivative $\frac{\partial Z^{[3]}_{2}}{\partial A^{[2]}_{1}}$ is the weight that connects $A^{[2]}_{1}$ to the $2^{nd}$ output node, i.e., $W^{[3]}_{21}$. The same applies to $E_{3}$:
	\[
	\frac{\partial E_{2}}{\partial W^{[2]}_{11}} = \delta^{[3]}_{2} \cdot W^{[3]}_{21} \cdot \sigma'\left(Z^{[2]}_{1}\right) \cdot A^{[1]}_{1},\;\;\;\;\;
	\frac{\partial E_{3}}{\partial W^{[2]}_{11}} = \delta^{[3]}_{3} \cdot W^{[3]}_{31} \cdot \sigma'\left(Z^{[2]}_{1}\right) \cdot A^{[1]}_{1}
	\]
	and by adding the three paths together and taking the common terms $\sigma'\left(Z^{[2]}_{1}\right)$ and $A^{[1]}_{1}$ out:
	\begin{equation}
	\begin{split}
	\frac{\partial E}{\partial W^{[2]}_{11}} &=
	\left(W^{[3]}_{11}\delta^{[3]}_{1} + W^{[3]}_{21}\delta^{[3]}_{2} + W^{[3]}_{31}\delta^{[3]}_{3}\right) \cdot \sigma'\left(Z^{[2]}_{1}\right) \cdot A^{[1]}_{1} \\
	&=
	\left(\sum_{p=1}^{3} W^{[3]}_{p1}\delta^{[3]}_{p}\right) \cdot \sigma'\left(Z^{[2]}_{1}\right) \cdot A^{[1]}_{1}
	\end{split}
	\label{dE_dW2_11_sum}
	\end{equation}
	In the same way as we did for the layer $3$, we will call everything before $A^{[1]}_{1}$ the \textbf{error} of the $1^{st}$ node of the layer $2$:
	\begin{equation}
	\boxed{
	\delta^{[2]}_{1} =
	\underbrace{
		\left(\sum_{p=1}^{3} W^{[3]}_{p1}\delta^{[3]}_{p}\right)
	}_{weighted\;propagated\;errors}
	\cdot
	\underbrace{
		\sigma'\left(Z^{[2]}_{1}\right)
	}_{activation's\;derivative}
	}
	\label{delta_2_1}
	\end{equation}
	\textbf{Note}: if the network had only one output node, the sum would have only one term and $\delta^{[2]}_{1}$ would be $\delta^{[3]}_{1}\cdot\frac{\partial Z^{[3]}_{1}}{\partial A^{[2]}_{1}}\cdot\frac{\partial A^{[2]}_{1}}{\partial Z^{[2]}_{1}}$, i.e., the terms of the single path in equation(\ref{chain_rule_layer2}). With many output nodes, we add up the errors of all output nodes, each one weighted by the weight that connects it to $A^{[2]}_{1}$. Again, $\delta^{[2]}_{1} = \frac{\partial E}{\partial Z^{[2]}_{1}}$.
	\\\\
	Hence,
	\begin{equation}
	\frac{\partial E}{\partial W^{[2]}_{11}} = \delta^{[2]}_{1} \cdot \frac{\partial Z^{[2]}_{1}}{\partial W^{[2]}_{11}} =
	\underbrace{\delta^{[2]}_{1}}_{propagated\;error}
	\cdot
	\underbrace{A^{[1]}_{1}}_{input\;to\;the\;layer}
	\label{dE_dW2_11}
	\end{equation}
	We can see clearly that equation(\ref{dE_dW2_11}) has exactly the same form as equation(\ref{dE_dW3_11}) of the output layer. The only difference is how the error $\delta$ is computed. For any weight $W^{[2]}_{jk}$ of the layer $2$, we will get:
	\begin{equation}
	\delta^{[2]}_{j} = \left(\sum_{p=1}^{3} W^{[3]}_{pj}\delta^{[3]}_{p}\right)\cdot\sigma'\left(Z^{[2]}_{j}\right),\;\;\;\;\;
	\frac{\partial E}{\partial W^{[2]}_{jk}} = \delta^{[2]}_{j} \cdot A^{[1]}_{k}
	\label{dE_dW2_jk}
	\end{equation}
	Let's write the errors of all the nodes of the layer $2$ in a single vector:
	\[
	\underbrace{\delta^{[2]}}_\text{$4\times 1$} =
	\underbrace{
		\begin{bmatrix}
		\delta^{[2]}_{1} \\
		\delta^{[2]}_{2} \\
		\delta^{[2]}_{3} \\
		\delta^{[2]}_{4}
		\end{bmatrix}}_\text{$4\times 1$}
	=
	\underbrace{
		\begin{bmatrix}
		W^{[3]}_{11}\delta^{[3]}_{1} + W^{[3]}_{21}\delta^{[3]}_{2} + W^{[3]}_{31}\delta^{[3]}_{3} \\
		W^{[3]}_{12}\delta^{[3]}_{1} + W^{[3]}_{22}\delta^{[3]}_{2} + W^{[3]}_{32}\delta^{[3]}_{3} \\
		W^{[3]}_{13}\delta^{[3]}_{1} + W^{[3]}_{23}\delta^{[3]}_{2} + W^{[3]}_{33}\delta^{[3]}_{3} \\
		W^{[3]}_{14}\delta^{[3]}_{1} + W^{[3]}_{24}\delta^{[3]}_{2} + W^{[3]}_{34}\delta^{[3]}_{3}
		\end{bmatrix}}_\text{$4\times 1$}
	\odot
	\underbrace{
		\begin{bmatrix}
		\sigma'\left(Z^{[2]}_{1}\right) \\
		\sigma'\left(Z^{[2]}_{2}\right) \\
		\sigma'\left(Z^{[2]}_{3}\right) \\
		\sigma'\left(Z^{[2]}_{4}\right)
		\end{bmatrix}}_\text{$4\times 1$}
	\]
	The first vector is a matrix multiplication. Notice that the $j^{th}$ row uses the $j^{th}$ \textbf{column} of $W^{[3]}$, which means that it is the $j^{th}$ \textbf{row} of the \textbf{transposed matrix} $W^{[3]T}$:
	\[
	\underbrace{
		\begin{bmatrix}
		W^{[3]}_{11}\delta^{[3]}_{1} + W^{[3]}_{21}\delta^{[3]}_{2} + W^{[3]}_{31}\delta^{[3]}_{3} \\
		W^{[3]}_{12}\delta^{[3]}_{1} + W^{[3]}_{22}\delta^{[3]}_{2} + W^{[3]}_{32}\delta^{[3]}_{3} \\
		W^{[3]}_{13}\delta^{[3]}_{1} + W^{[3]}_{23}\delta^{[3]}_{2} + W^{[3]}_{33}\delta^{[3]}_{3} \\
		W^{[3]}_{14}\delta^{[3]}_{1} + W^{[3]}_{24}\delta^{[3]}_{2} + W^{[3]}_{34}\delta^{[3]}_{3}
		\end{bmatrix}}_\text{$4\times 1$}
	=
	\underbrace{
    \underbrace{
		\begin{bmatrix}
		W^{[3]}_{11} & W^{[3]}_{21} & W^{[3]}_{31} \\
		W^{[3]}_{12} & W^{[3]}_{22} & W^{[3]}_{32} \\
		W^{[3]}_{13} & W^{[3]}_{23} & W^{[3]}_{33} \\
		W^{[3]}_{14} & W^{[3]}_{24} & W^{[3]}_{34}
		\end{bmatrix}}_\text{$4\times 3$}
	\cdot
	\underbrace{
		\begin{bmatrix}
		\delta^{[3]}_{1} \\
		\delta^{[3]}_{2} \\
		\delta^{[3]}_{3}
		\end{bmatrix}}_\text{$3\times 1$}}_\text{$4\times 1$}
	=
	\underbrace{\underbrace{W^{[3]T}}_\text{$4\times 3$} \cdot \underbrace{\delta^{[3]}}_\text{$3\times 1$}}_\text{$4\times 1$}
	\]
	hence,
	\begin{equation}
	\underbrace{\delta^{[2]}}_\text{$4\times 1$} =
	\underbrace{
		\left(\underbrace{W^{[3]T}}_\text{$4\times 3$} \cdot \underbrace{\delta^{[3]}}_\text{$3\times 1$}\right)
	}_\text{$4\times 1$}
	\odot
	\underbrace{\sigma'\left(Z^{[2]}\right)}_\text{$4\times 1$}
	\label{delta_2_vector}
	\end{equation}
	and by using the same outer product steps that we used to get equation(\ref{dE_dW3}) and equation(\ref{dE_db3}):
	\begin{equation}
	\underbrace{\frac{\partial E}{\partial W^{[2]}}}_\text{$4\times 5$} = \underbrace{\underbrace{\delta^{[2]}}_\text{$4\times 1$} \cdot \underbrace{A^{[1]T}}_\text{$1\times 5$}}_\text{$4\times 5$},\;\;\;\;\;
	\underbrace{\frac{\partial E}{\partial b^{[2]}}}_\text{$4\times 1$} = \underbrace{\delta^{[2]}}_\text{$4\times 1$}
	\label{dE_dW2}
	\end{equation}
	The dimensions $4\times 5$ match the dimensions of $W^{[2]}$ as expected.

	\subsection{The First Hidden Layer (Layer 1)}
	Finally, let's take the first weight on the bold path in Figure(\ref{network_figure}), $W^{[1]}_{11}$. If we follow the bold path to $E_{1}$, the chain rule gets longer:
	\[
	W^{[1]}_{11} \longrightarrow Z^{[1]}_{1} \longrightarrow A^{[1]}_{1} \longrightarrow Z^{[2]}_{1} \longrightarrow A^{[2]}_{1} \longrightarrow Z^{[3]}_{1} \longrightarrow \hat{y}_{1} \longrightarrow E_{1}
	\]
	\begin{equation}
	\frac{\partial E_{1}}{\partial W^{[1]}_{11}} =
	\underbrace{
		\underbrace{
			\underbrace{
				\frac{\partial E_{1}}{\partial \hat{y}_{1}} \cdot
				\frac{\partial \hat{y}_{1}}{\partial Z^{[3]}_{1}}
			}_{\delta^{[3]}_{1}}
			\cdot
			\frac{\partial Z^{[3]}_{1}}{\partial A^{[2]}_{1}} \cdot
			\frac{\partial A^{[2]}_{1}}{\partial Z^{[2]}_{1}}
		}_{\delta^{[2]}_{1}}
		\cdot
		\frac{\partial Z^{[2]}_{1}}{\partial A^{[1]}_{1}} \cdot
		\frac{\partial A^{[1]}_{1}}{\partial Z^{[1]}_{1}}
	}_{\delta^{[1]}_{1}}
	\cdot
	\frac{\partial Z^{[1]}_{1}}{\partial W^{[1]}_{11}}
	\label{chain_rule_layer1}
	\end{equation}
	We can see the pattern: each time we move one layer backward, we multiply the error of the next layer by a \textbf{weight} and an \textbf{activation's derivative}. As we saw in the layer $2$, the braces above are the single-path versions of the errors. To get the complete derivative, we have to add up all the paths.
	\\\\
	The output $A^{[1]}_{1}$ goes to all the $4$ nodes of the layer $2$, and each of them goes to all the $3$ output nodes, so there are $4\times 3 = 12$ paths from $W^{[1]}_{11}$ to the error $E$. Fortunately, we do not need to follow all of them. Recall that $\delta^{[2]}_{p} = \frac{\partial E}{\partial Z^{[2]}_{p}}$ already contains the sum over all the paths from $Z^{[2]}_{p}$ to the error. Therefore, we only need to add up the $4$ paths that go from $A^{[1]}_{1}$ to the layer $2$:
	\[
	\frac{\partial E}{\partial W^{[1]}_{11}} =
	\sum_{p=1}^{4}
	\underbrace{
		\frac{\partial E}{\partial Z^{[2]}_{p}}
	}_{\delta^{[2]}_{p}}
	\cdot
	\underbrace{
		\frac{\partial Z^{[2]}_{p}}{\partial A^{[1]}_{1}}
	}_{W^{[2]}_{p1}}
	\cdot
	\frac{\partial A^{[1]}_{1}}{\partial Z^{[1]}_{1}}
	\cdot
	\frac{\partial Z^{[1]}_{1}}{\partial W^{[1]}_{11}}
	\]
	The last two terms are the same for all the paths, $\sigma'\left(Z^{[1]}_{1}\right)$ and $A^{[0]}_{1}$, so we can take them out of the sum:
	\begin{equation}
	\frac{\partial E}{\partial W^{[1]}_{11}} =
	\underbrace{
		\left(\sum_{p=1}^{4} W^{[2]}_{p1}\delta^{[2]}_{p}\right)\cdot\sigma'\left(Z^{[1]}_{1}\right)
	}_{\delta^{[1]}_{1}}
	\cdot\; A^{[0]}_{1}
	= \delta^{[1]}_{1}\cdot A^{[0]}_{1}
	\label{dE_dW1_11}
	\end{equation}
	This is the main idea of backpropagation: the errors $\delta$ of a layer are computed \textbf{once} and then they are \textbf{reused} to compute the errors of the previous layer. The errors are propagated backward one layer at a time.
	\\\\
	By following the same steps that we used to get equation(\ref{delta_2_vector}) and equation(\ref{dE_dW2}), we will get:
	\begin{equation}
	\underbrace{\delta^{[1]}}_\text{$5\times 1$} =
	\underbrace{
		\left(\underbrace{W^{[2]T}}_\text{$5\times 4$} \cdot \underbrace{\delta^{[2]}}_\text{$4\times 1$}\right)
	}_\text{$5\times 1$}
	\odot
	\underbrace{\sigma'\left(Z^{[1]}\right)}_\text{$5\times 1$}
	\label{delta_1_vector}
	\end{equation}
	\begin{equation}
	\underbrace{\frac{\partial E}{\partial W^{[1]}}}_\text{$5\times 3$} = \underbrace{\underbrace{\delta^{[1]}}_\text{$5\times 1$} \cdot \underbrace{A^{[0]T}}_\text{$1\times 3$}}_\text{$5\times 3$} = \underbrace{\underbrace{\delta^{[1]}}_\text{$5\times 1$} \cdot \underbrace{x^{T}}_\text{$1\times 3$}}_\text{$5\times 3$},\;\;\;\;\;
	\underbrace{\frac{\partial E}{\partial b^{[1]}}}_\text{$5\times 1$} = \underbrace{\delta^{[1]}}_\text{$5\times 1$}
	\label{dE_dW1}
	\end{equation}

	\subsection{The General Backpropagation Equations}
	If we compare the results of the three layers, we can write the general equations of backpropagation for any network with $L$ layers. The error of the \textbf{last layer} comes from equation(\ref{delta_3_vector}), where $\hat{y} = A^{[L]}$ is the output of the last layer:
	\begin{equation}
	\boxed{
	\underbrace{\delta^{[L]}}_\text{$n_{L}\times 1$} =
	\overbrace{
		\underbrace{-\left(y - \hat{y}\right)}_\text{$n_{L}\times 1$}
	}^{loss\;derivative}
	\odot
	\overbrace{
		\underbrace{\sigma'\left(Z^{[L]}\right)}_\text{$n_{L}\times 1$}
	}^{activation's\;derivative}
	}
	\label{general_delta_L}
	\end{equation}
	The error of any \textbf{hidden layer} $l = L-1, L-2, \dots, 1$ comes from equations(\ref{delta_2_vector}) and (\ref{delta_1_vector}):
	\begin{equation}
	\boxed{
	\underbrace{\delta^{[l]}}_\text{$n_{l}\times 1$} =
	\overbrace{
		\underbrace{
			\left(\underbrace{W^{[l+1]T}}_\text{$n_{l}\times n_{l+1}$} \cdot \underbrace{\delta^{[l+1]}}_\text{$n_{l+1}\times 1$}\right)
		}_\text{$n_{l}\times 1$}
	}^{weighted\;propagated\;errors\;from\;the\;next\;layer}
	\odot
	\overbrace{
		\underbrace{\sigma'\left(Z^{[l]}\right)}_\text{$n_{l}\times 1$}
	}^{activation's\;derivative}
	}
	\label{general_delta_l}
	\end{equation}
	The partial derivatives of the weights and the biases of any layer $l = L, L-1, \dots, 1$ come from equations(\ref{dE_dW3}), (\ref{dE_db3}), (\ref{dE_dW2}), and (\ref{dE_dW1}):
	\begin{equation}
	\boxed{
	\underbrace{\frac{\partial E}{\partial W^{[l]}}}_\text{$n_{l}\times n_{l-1}$} =
    \underbrace{
    \overbrace{
		\underbrace{\delta^{[l]}}_\text{$n_{l}\times 1$}
	}^{propagated\;error}
	\cdot
	\overbrace{
		\underbrace{A^{[l-1]T}}_\text{$1\times n_{l-1}$}
	}^{input\;to\;the\;layer}}_\text{$n_l\times n_{l-1}$}
	}
	\label{general_dE_dW}
	\end{equation}
	\begin{equation}
	\boxed{
	\underbrace{\frac{\partial E}{\partial b^{[l]}}}_\text{$n_{l}\times 1$} = \underbrace{\delta^{[l]}}_\text{$n_{l}\times 1$}
	}
	\label{general_dE_db}
	\end{equation}
	\textbf{Note}: equation(\ref{general_delta_l}) uses the \textbf{transposed} weights matrix of the \textbf{next} layer. In the forward propagation, $W^{[l+1]}$ carries the outputs forward from the layer $l$ to the layer $(l+1)$. In the backward propagation, $W^{[l+1]T}$ carries the errors backward from the layer $(l+1)$ to the layer $l$ through the same connections.
	\\\\
	\textbf{Note}: if we write the errors as row-vectors instead of column-vectors, i.e., $\delta^{[l]T}$, then by taking the transpose of both sides of equation(\ref{general_delta_l}), it becomes $\delta^{[l]T} = \left(\delta^{[l+1]T}\cdot W^{[l+1]}\right) \odot \sigma'\left(Z^{[l]}\right)^{T}$. Both forms are the same equation.

	\subsection{Backpropagation over $m$ Examples}
	\label{batch_section}
	So far, we have derived the partial derivatives of the error $E$ of a single example. However, the cost function $J$ in equation(\ref{cost_function}) is the average over $m$ examples. Since the derivative of a sum is the sum of the derivatives:
	\begin{equation}
	\frac{\partial J}{\partial W^{[l]}} = \frac{1}{m}\sum_{i=1}^{m}\frac{\partial E^{(i)}}{\partial W^{[l]}} = \frac{1}{m}\sum_{i=1}^{m} \underbrace{\delta^{[l](i)}}_\text{$n_{l}\times 1$} \cdot \underbrace{A^{[l-1](i)T}}_\text{$1\times n_{l-1}$}
	\label{dJ_dW_sum}
	\end{equation}
	where $\delta^{[l](i)}$ and $A^{[l-1](i)}$ are the error and the input of the layer $l$ when the $i^{th}$ example is fed to the network. Instead of running the network $m$ times with a loop, we can process all the examples at once by stacking them as \textbf{columns} next to each other. We will use the \textbf{bold} letters $\mathbf{A}^{[l]}$ and $\mathbf{Z}^{[l]}$ for the stacked versions of $A^{[l]}$ and $Z^{[l]}$, the \textbf{bold} capital letter $\boldsymbol{\Delta}^{[l]}$ for the stacked version of $\delta^{[l]}$, and the capital letters $X$, $Y$, and $\hat{Y}$ for the stacked versions of $x$, $y$, and $\hat{y}$:
	\[
	\underbrace{X}_\text{$n_{0}\times m$} =
	\underbrace{
		\begin{bmatrix}
		x^{(1)}_{1} & x^{(2)}_{1} & x^{(3)}_{1} & \dots  & x^{(m)}_{1} \\
		x^{(1)}_{2} & x^{(2)}_{2} & x^{(3)}_{2} & \dots  & x^{(m)}_{2} \\
		\vdots      & \vdots      & \vdots      & \ddots & \vdots      \\
		x^{(1)}_{n_{0}} & x^{(2)}_{n_{0}} & x^{(3)}_{n_{0}} & \dots  & x^{(m)}_{n_{0}}
		\end{bmatrix}}_\text{$n_{0}\times m$}
	\]
	\[
	\underbrace{\mathbf{A}^{[l]}}_\text{$n_{l}\times m$} =
	\underbrace{
		\begin{bmatrix}
		|           & |           &        & |           \\
		A^{[l](1)}  & A^{[l](2)}  & \dots  & A^{[l](m)}  \\
		|           & |           &        & |
		\end{bmatrix}}_\text{$n_{l}\times m$}
	\]
	\[
	\underbrace{Y}_\text{$n_{L}\times m$} =
	\underbrace{
		\begin{bmatrix}
		|           & |           &        & |           \\
		y^{(1)}     & y^{(2)}     & \dots  & y^{(m)}     \\
		|           & |           &        & |
		\end{bmatrix}}_\text{$n_{L}\times m$},\;\;\;\;\;
	\underbrace{\boldsymbol{\Delta}^{[l]}}_\text{$n_{l}\times m$} =
	\underbrace{
		\begin{bmatrix}
		|              & |              &        & |              \\
		\delta^{[l](1)} & \delta^{[l](2)} & \dots  & \delta^{[l](m)} \\
		|              & |              &        & |
		\end{bmatrix}}_\text{$n_{l}\times m$}
	\]
	The predicted values of all the examples are the outputs of the last layer:
	\[
	\underbrace{\hat{Y}}_\text{$n_{L}\times m$} =
	\underbrace{
		\begin{bmatrix}
		|                & |                &        & |                \\
		\hat{y}^{(1)}    & \hat{y}^{(2)}    & \dots  & \hat{y}^{(m)}    \\
		|                & |                &        & |
		\end{bmatrix}}_\text{$n_{L}\times m$}
	=
	\underbrace{\mathbf{A}^{[L]}}_\text{$n_{L}\times m$}
	\]
	and $\mathbf{Z}^{[l]}$ is stacked in the same way as $\mathbf{A}^{[l]}$. The $i^{th}$ column of each matrix belongs to the $i^{th}$ example.
	\\\\
	For the forward propagation, if we multiply $W^{[l]}$ by $\mathbf{A}^{[l-1]}$, each column of $\mathbf{A}^{[l-1]}$ is multiplied by $W^{[l]}$ separately. The bias vector must be added to every column, so we multiply it by a row-vector of ones $\mathbf{1}^{T}$, where $\mathbf{1}$ is an $m\times 1$ vector of ones:
	\begin{equation}
	\begin{split}
	\underbrace{\mathbf{A}^{[0]}}_\text{$n_{0}\times m$} &= \underbrace{X}_\text{$n_{0}\times m$} \\
	\underbrace{\mathbf{Z}^{[l]}}_\text{$n_{l}\times m$} &= \underbrace{\underbrace{W^{[l]}}_\text{$n_{l}\times n_{l-1}$} \cdot \underbrace{\mathbf{A}^{[l-1]}}_\text{$n_{l-1}\times m$}}_\text{$n_l\times m$} + \underbrace{\underbrace{b^{[l]}}_\text{$n_{l}\times 1$} \cdot \underbrace{\mathbf{1}^{T}}_\text{$1\times m$}}_\text{$n_l\times m$},\;\;\;\;\;
	\underbrace{\mathbf{A}^{[l]}}_\text{$n_{l}\times m$} = \sigma\left(\underbrace{\mathbf{Z}^{[l]}}_\text{$n_{l}\times m$}\right)
	\end{split}
	\label{forward_propagation_batch}
	\end{equation}
	where
	\[
	\underbrace{\underbrace{b^{[l]}}_\text{$n_{l}\times 1$} \cdot \underbrace{\mathbf{1}^{T}}_\text{$1\times m$}}_\text{$n_l\times m$} =
	\underbrace{\underbrace{
		\begin{bmatrix}
		b^{[l]}_{1} \\
		b^{[l]}_{2} \\
		\vdots      \\
		b^{[l]}_{n_{l}}
		\end{bmatrix}}_\text{$n_{l}\times 1$}
	\cdot
	\underbrace{
		\begin{bmatrix}
		1 & 1 & \dots & 1
		\end{bmatrix}}_\text{$1\times m$}}_\text{$n_l\times m$}
	=
	\underbrace{
		\begin{bmatrix}
		b^{[l]}_{1} & b^{[l]}_{1} & \dots  & b^{[l]}_{1} \\
		b^{[l]}_{2} & b^{[l]}_{2} & \dots  & b^{[l]}_{2} \\
		\vdots      & \vdots      & \ddots & \vdots      \\
		b^{[l]}_{n_{l}} & b^{[l]}_{n_{l}} & \dots  & b^{[l]}_{n_{l}}
		\end{bmatrix}}_\text{$n_{l}\times m$}
	\]
	\textbf{Note}: the product $b^{[l]}\cdot\mathbf{1}^{T}$ copies the bias vector $m$ times, once for every example. This is exactly what is called \textbf{broadcasting} in NumPy and PyTorch. In code, we simply write \texttt{Z = W.dot(A\_prev) + b}, where \texttt{b} has the shape $(n_{l}, 1)$, and the library adds \texttt{b} to every column of \texttt{W.dot(A\_prev)} without building the $n_{l}\times m$ matrix in memory. The bias must be stored as a column-vector of shape $(n_{l}, 1)$. If it is stored as a 1-D array of shape $(n_{l},)$, NumPy matches it with the \textbf{last} dimension of the matrix, which is the number of examples $m$, not the number of nodes $n_{l}$.
	\\\\
	The backward pass does the opposite of broadcasting. Since the forward pass copies $b^{[l]}$ into all the $m$ columns, the gradient of $b^{[l]}$ \textbf{adds up} the $m$ columns of the errors. We will see this in equation(\ref{dJ_dW_batch}), where $\boldsymbol{\Delta}^{[l]}\cdot\mathbf{1}$ is the same as \texttt{delta.sum(axis=1, keepdims=True)} in NumPy.
	\\\\
	The equations(\ref{general_delta_L}) and (\ref{general_delta_l}) work on each column separately too, since the matrix product and the Hadamard product do not mix the columns. Hence,
	\begin{equation}
	\begin{split}
	\underbrace{\boldsymbol{\Delta}^{[L]}}_\text{$n_{L}\times m$} &= \underbrace{-\left(Y - \hat{Y}\right)}_\text{$n_{L}\times m$} \odot \underbrace{\sigma'\left(\mathbf{Z}^{[L]}\right)}_\text{$n_{L}\times m$} \\
	\underbrace{\boldsymbol{\Delta}^{[l]}}_\text{$n_{l}\times m$} &= \underbrace{\left(\underbrace{W^{[l+1]T}}_\text{$n_{l}\times n_{l+1}$} \cdot \underbrace{\boldsymbol{\Delta}^{[l+1]}}_\text{$n_{l+1}\times m$}\right)}_\text{$n_{l}\times m$} \odot \underbrace{\sigma'\left(\mathbf{Z}^{[l]}\right)}_\text{$n_{l}\times m$}
	\end{split}
	\label{delta_batch}
	\end{equation}
	For the weights, \textbf{the sum of outer products} in equation(\ref{dJ_dW_sum}) is exactly a \textbf{matrix multiplication}. If we multiply $\boldsymbol{\Delta}^{[l]}$ by $\mathbf{A}^{[l-1]T}$, each column of the first matrix is multiplied by the corresponding row of the second one and the results are added up:
	\[
	\underbrace{\boldsymbol{\Delta}^{[l]}}_\text{$n_{l}\times m$} \cdot \underbrace{\mathbf{A}^{[l-1]T}}_\text{$m\times n_{l-1}$}
	=
	\underbrace{
		\begin{bmatrix}
		|              & |              &        & |              \\
		\delta^{[l](1)} & \delta^{[l](2)} & \dots  & \delta^{[l](m)} \\
		|              & |              &        & |
		\end{bmatrix}}_\text{$n_{l}\times m$}
	\cdot
	\underbrace{
		\begin{bmatrix}
		- & A^{[l-1](1)T} & - \\
		- & A^{[l-1](2)T} & - \\
		  & \vdots        &   \\
		- & A^{[l-1](m)T} & -
		\end{bmatrix}}_\text{$m\times n_{l-1}$}
	\]
	\[
	=
	\underbrace{
		\sum_{i=1}^{m} \delta^{[l](i)} \cdot A^{[l-1](i)T}
	}_\text{$n_{l}\times n_{l-1}$}
	\]
	By plugging this result into equation(\ref{dJ_dW_sum}), and by following the same steps for the biases (adding up the columns of $\boldsymbol{\Delta}^{[l]}$ is the same as multiplying it by $\mathbf{1}$), we will get:
	\begin{equation}
	\boxed{
	\underbrace{\frac{\partial J}{\partial W^{[l]}}}_\text{$n_{l}\times n_{l-1}$} = \frac{1}{m}\cdot\underbrace{\underbrace{\boldsymbol{\Delta}^{[l]}}_\text{$n_{l}\times m$} \cdot \underbrace{\mathbf{A}^{[l-1]T}}_\text{$m\times n_{l-1}$}}_\text{$n_l\times n_{l-1}$},\;\;\;\;\;
	\underbrace{\frac{\partial J}{\partial b^{[l]}}}_\text{$n_{l}\times 1$} = \frac{1}{m}\cdot\underbrace{\underbrace{\boldsymbol{\Delta}^{[l]}}_\text{$n_{l}\times m$} \cdot \underbrace{\mathbf{1}}_\text{$m\times 1$}}_\text{$n_l\times 1$}
	}
	\label{dJ_dW_batch}
	\end{equation}

	\subsection{Updating The Weights (Gradient Descent)}
	Gradient descent updates the weights with the following \textbf{update rule}: $w_{new} = w_{old} - \alpha\cdot\nabla_{w}{RSS}$, where $\alpha$ is the learning rate. We showed how to derive this rule using the Taylor series in the linear regression paper~\cite{alasmary_linear_regression}, and the reader can refer to it for more details. The same rule is applied here to every weight and every bias. For a single example, the update of the first column of $W^{[1]}$ is shown below, where the arrow $\leftarrow$ means that the old value is replaced by the new one:
	\[
	\begin{split}
	W^{[1]}_{11} &\leftarrow W^{[1]}_{11} - \alpha\cdot\delta^{[1]}_{1}\cdot A^{[0]}_{1} \\
	W^{[1]}_{21} &\leftarrow W^{[1]}_{21} - \alpha\cdot\delta^{[1]}_{2}\cdot A^{[0]}_{1} \\
	W^{[1]}_{31} &\leftarrow W^{[1]}_{31} - \alpha\cdot\delta^{[1]}_{3}\cdot A^{[0]}_{1} \\
	W^{[1]}_{41} &\leftarrow W^{[1]}_{41} - \alpha\cdot\delta^{[1]}_{4}\cdot A^{[0]}_{1} \\
	W^{[1]}_{51} &\leftarrow W^{[1]}_{51} - \alpha\cdot\delta^{[1]}_{5}\cdot A^{[0]}_{1}
	\end{split}
	\]
	and by doing the same for all the weights of the layer $1$ at once, using equation(\ref{dE_dW1}):
	\[
	\underbrace{W^{[1]}}_\text{$5\times 3$} - \alpha\cdot\underbrace{\underbrace{\delta^{[1]}}_\text{$5\times 1$}\cdot\underbrace{A^{[0]T}}_\text{$1\times 3$}}_\text{$5\times 3$}
	=
	\underbrace{
		\begin{bmatrix}
		W^{[1]}_{11} & W^{[1]}_{12} & W^{[1]}_{13} \\
		W^{[1]}_{21} & W^{[1]}_{22} & W^{[1]}_{23} \\
		W^{[1]}_{31} & W^{[1]}_{32} & W^{[1]}_{33} \\
		W^{[1]}_{41} & W^{[1]}_{42} & W^{[1]}_{43} \\
		W^{[1]}_{51} & W^{[1]}_{52} & W^{[1]}_{53}
		\end{bmatrix}}_\text{$5\times 3$}
	- \alpha\cdot
	\underbrace{
		\begin{bmatrix}
		\delta^{[1]}_{1}A^{[0]}_{1} & \delta^{[1]}_{1}A^{[0]}_{2} & \delta^{[1]}_{1}A^{[0]}_{3} \\
		\delta^{[1]}_{2}A^{[0]}_{1} & \delta^{[1]}_{2}A^{[0]}_{2} & \delta^{[1]}_{2}A^{[0]}_{3} \\
		\delta^{[1]}_{3}A^{[0]}_{1} & \delta^{[1]}_{3}A^{[0]}_{2} & \delta^{[1]}_{3}A^{[0]}_{3} \\
		\delta^{[1]}_{4}A^{[0]}_{1} & \delta^{[1]}_{4}A^{[0]}_{2} & \delta^{[1]}_{4}A^{[0]}_{3} \\
		\delta^{[1]}_{5}A^{[0]}_{1} & \delta^{[1]}_{5}A^{[0]}_{2} & \delta^{[1]}_{5}A^{[0]}_{3}
		\end{bmatrix}}_\text{$5\times 3$}
	\]
	\[
	=
	\underbrace{
		\begin{bmatrix}
		W^{[1]}_{11} - \alpha\,\delta^{[1]}_{1}A^{[0]}_{1} & W^{[1]}_{12} - \alpha\,\delta^{[1]}_{1}A^{[0]}_{2} & W^{[1]}_{13} - \alpha\,\delta^{[1]}_{1}A^{[0]}_{3} \\
		W^{[1]}_{21} - \alpha\,\delta^{[1]}_{2}A^{[0]}_{1} & W^{[1]}_{22} - \alpha\,\delta^{[1]}_{2}A^{[0]}_{2} & W^{[1]}_{23} - \alpha\,\delta^{[1]}_{2}A^{[0]}_{3} \\
		W^{[1]}_{31} - \alpha\,\delta^{[1]}_{3}A^{[0]}_{1} & W^{[1]}_{32} - \alpha\,\delta^{[1]}_{3}A^{[0]}_{2} & W^{[1]}_{33} - \alpha\,\delta^{[1]}_{3}A^{[0]}_{3} \\
		W^{[1]}_{41} - \alpha\,\delta^{[1]}_{4}A^{[0]}_{1} & W^{[1]}_{42} - \alpha\,\delta^{[1]}_{4}A^{[0]}_{2} & W^{[1]}_{43} - \alpha\,\delta^{[1]}_{4}A^{[0]}_{3} \\
		W^{[1]}_{51} - \alpha\,\delta^{[1]}_{5}A^{[0]}_{1} & W^{[1]}_{52} - \alpha\,\delta^{[1]}_{5}A^{[0]}_{2} & W^{[1]}_{53} - \alpha\,\delta^{[1]}_{5}A^{[0]}_{3}
		\end{bmatrix}}_\text{$5\times 3$}
	\]
	For $m$ examples, we use the gradients of the cost $J$ from equation(\ref{dJ_dW_batch}):
	\begin{equation}
	\begin{split}
	W^{[l]}_{new} &= W^{[l]}_{old} - \alpha\cdot\frac{\partial J}{\partial W^{[l]}} \\
	b^{[l]}_{new} &= b^{[l]}_{old} - \alpha\cdot\frac{\partial J}{\partial b^{[l]}}
	\end{split}
	\label{update_rule}
	\end{equation}
	\textbf{Note}: all the errors $\boldsymbol{\Delta}^{[l]}$ must be computed with the \textbf{old} weights. In Algorithm(\ref{backprop_algorithm}), we compute all the gradients first and update the weights only after the backward pass is finished. In the algorithm, we write $\nabla_{W^{[l]}}$ and $\nabla_{b^{[l]}}$ for the gradients $\frac{\partial J}{\partial W^{[l]}}$ and $\frac{\partial J}{\partial b^{[l]}}$.
	\\\\
	\textbf{Note}: one pass over all the $m$ examples of the dataset is called an \textbf{epoch}. In Algorithm(\ref{backprop_algorithm}), every epoch computes the gradients of all the examples and then updates the weights only one time. The number of epochs $T$ is the number of times we repeat this.

	\begin{algorithm}[htbp]
		\caption{Batch Gradient Descent with Backpropagation}
		\label{backprop_algorithm}
		\begin{algorithmic}[1]
			\State $W^{[l]}$ = a randomly initialized matrix of size $n_{l} \times n_{l-1}$, for $l = 1, \dots, L$ \label{alg1:init_W}
			\State $b^{[l]}$ = a vector of zeros of size $n_{l} \times 1$, for $l = 1, \dots, L$ \label{alg1:init_b}
			\State $\alpha$ = the learning rate
			\State $T$ = the number of epochs \label{alg1:epochs}
			\For {$e = 1$ to $T$} \label{alg1:epoch_loop}
				\State \textit{// forward propagation} \label{alg1:forward_start}
				\State $\mathbf{A}^{[0]} = X$ \label{alg1:A0}
				\For {$l = 1$ to $L$}
					\State $\mathbf{Z}^{[l]} = W^{[l]} \cdot \mathbf{A}^{[l-1]} + b^{[l]} \cdot \mathbf{1}^{T}$ \label{alg1:Z}
					\State $\mathbf{A}^{[l]} = \sigma\left(\mathbf{Z}^{[l]}\right)$ \label{alg1:A}
				\EndFor
				\State $\hat{Y} = \mathbf{A}^{[L]}$ \label{alg1:Yhat}
				\State \textit{// backward propagation}
				\State $\boldsymbol{\Delta}^{[L]} = -\left(Y - \hat{Y}\right) \odot \sigma'\left(\mathbf{Z}^{[L]}\right)$ \label{alg1:delta_L}
				\For {$l = L$ down to $1$}
					\State $\nabla_{W^{[l]}} = \frac{1}{m}\cdot\boldsymbol{\Delta}^{[l]} \cdot \mathbf{A}^{[l-1]T}$ \label{alg1:grad_W}
					\State $\nabla_{b^{[l]}} = \frac{1}{m}\cdot\boldsymbol{\Delta}^{[l]} \cdot \mathbf{1}$ \label{alg1:grad_b}
					\If {$l > 1$}
						\State $\boldsymbol{\Delta}^{[l-1]} = \left(W^{[l]T} \cdot \boldsymbol{\Delta}^{[l]}\right) \odot \sigma'\left(\mathbf{Z}^{[l-1]}\right)$ \label{alg1:delta_l}
					\EndIf
				\EndFor
				\State \textit{// update} \label{alg1:update_start}
				\For {$l = 1$ to $L$}
					\State $W^{[l]} \leftarrow W^{[l]} - \alpha \cdot \nabla_{W^{[l]}}$ \label{alg1:update_W}
					\State $b^{[l]} \leftarrow b^{[l]} - \alpha \cdot \nabla_{b^{[l]}}$ \label{alg1:update_b}
				\EndFor \label{alg1:update_end}
			\EndFor
		\end{algorithmic}
	\end{algorithm}

	\newpage
	\section{Mini-Batch Gradient Descent}
	\label{minibatch_section}
	In Algorithm(\ref{backprop_algorithm}), every update of the weights uses \textbf{all} the $m$ examples of the dataset. We compute the errors $\boldsymbol{\Delta}^{[l]}$ of all the examples, and only then we take a single step. This is called \textbf{batch gradient descent}. Since one \textbf{epoch} is one pass over all the examples, as we noted before Algorithm(\ref{backprop_algorithm}), batch gradient descent makes only \textbf{one update per epoch}. When the number of examples $m$ is large, this causes two problems:
	\begin{enumerate}[leftmargin=1cm, label=\arabic*.]
		\item Every single step needs a forward pass and a backward pass over the whole dataset, so the training is slow.
		\item The matrices $\mathbf{Z}^{[l]}$, $\mathbf{A}^{[l]}$, and $\boldsymbol{\Delta}^{[l]}$ have $m$ columns, and all of them must be stored in memory at the same time. For example, the dataset that we will use in Section(\ref{code_section}) has $m = 60,000$ examples.
	\end{enumerate}
	The other extreme is to update the weights after \textbf{every single example}. This is called \textbf{stochastic gradient descent (SGD)}. Each update is very cheap, but it also causes two problems. First, the gradient of a single example is \textbf{noisy}: it shows how to reduce the error of that example only, which may increase the errors of the other examples. Second, we lose the speed of the matrix multiplications, since we process one column at a time instead of many columns at once.
	\\\\
	\textbf{Mini-batch gradient descent} is the middle ground between the two. We split the dataset into small groups of examples called \textbf{mini-batches}, and we update the weights after every mini-batch.

	\subsection{Splitting The Dataset into Mini-Batches}
	Let $B$ be the \textbf{mini-batch size}, i.e., the number of examples in each mini-batch. Before splitting, we \textbf{shuffle} the columns of $X$ and $Y$ randomly, using the same random order for both matrices so that every example stays with its target values. Shuffling makes every mini-batch a random sample of the dataset. Without it, if the dataset is sorted by the labels (e.g., all the images of the digit $0$ first), a mini-batch would contain the examples of a single class only, and its gradient would push the network towards that class.
	\\\\
	After shuffling, we split the columns of $X$ and $Y$ into $N$ mini-batches:
	\[
	\underbrace{X}_\text{$n_{0}\times m$} =
	\Big[\;
	\underbrace{X^{\{1\}}}_\text{$n_{0}\times m_{1}$} \;\Big|\;
	\underbrace{X^{\{2\}}}_\text{$n_{0}\times m_{2}$} \;\Big|\; \dots \;\Big|\;
	\underbrace{X^{\{N\}}}_\text{$n_{0}\times m_{N}$}
	\;\Big]
	\]
	\[
	\underbrace{Y}_\text{$n_{L}\times m$} =
	\Big[\;
	\underbrace{Y^{\{1\}}}_\text{$n_{L}\times m_{1}$} \;\Big|\;
	\underbrace{Y^{\{2\}}}_\text{$n_{L}\times m_{2}$} \;\Big|\; \dots \;\Big|\;
	\underbrace{Y^{\{N\}}}_\text{$n_{L}\times m_{N}$}
	\;\Big]
	\]
	where $X^{\{t\}}$ and $Y^{\{t\}}$ are the features and the target values of the $t^{th}$ mini-batch, and $m_{t}$ is its number of examples. The number of mini-batches is:
	\begin{equation}
	N = \left\lceil \frac{m}{B} \right\rceil
	\label{number_of_minibatches}
	\end{equation}
	where $\lceil\;\rceil$ rounds up to the nearest integer. All the mini-batches have $m_{t} = B$ examples, except the last one which has the remaining $m_{N} = m - (N-1)\cdot B$ examples if $m$ is not divisible by $B$. For example, if $m = 10$ and $B = 4$, then $N = \lceil 10/4 \rceil = 3$ and the three mini-batches are:
	\[
	\underbrace{X^{\{1\}}}_\text{$n_{0}\times 4$} =
	\begin{bmatrix}
	|       & |       & |       & |       \\
	x^{(1)} & x^{(2)} & x^{(3)} & x^{(4)} \\
	|       & |       & |       & |
	\end{bmatrix},\;\;\;
	\underbrace{X^{\{2\}}}_\text{$n_{0}\times 4$} =
	\begin{bmatrix}
	|       & |       & |       & |       \\
	x^{(5)} & x^{(6)} & x^{(7)} & x^{(8)} \\
	|       & |       & |       & |
	\end{bmatrix}
	\]
	\[
	\underbrace{X^{\{3\}}}_\text{$n_{0}\times 2$} =
	\begin{bmatrix}
	|       & |        \\
	x^{(9)} & x^{(10)} \\
	|       & |
	\end{bmatrix}
	\]
	i.e., $m_{1} = m_{2} = 4$ and $m_{3} = 10 - 2\cdot 4 = 2$ (here the examples are numbered after shuffling).

	\subsection{The Cost and The Gradients of a Mini-Batch}
	The cost of the $t^{th}$ mini-batch is the average of the errors of its examples:
	\begin{equation}
	J^{\{t\}} = \frac{1}{m_{t}}\sum_{i\,\in\,\{t\}} E^{(i)}
	\label{minibatch_cost}
	\end{equation}
	where $i \in \{t\}$ means all the examples $i$ that belong to the $t^{th}$ mini-batch. A mini-batch is simply a smaller dataset. Therefore, all the equations of Section(\ref{batch_section}) still hold if we replace $X$ by $X^{\{t\}}$, $Y$ by $Y^{\{t\}}$, and $m$ by $m_{t}$. The forward propagation in equation(\ref{forward_propagation_batch}) becomes:
	\begin{equation}
	\begin{split}
	\underbrace{\mathbf{A}^{[0]\{t\}}}_\text{$n_{0}\times m_{t}$} &= \underbrace{X^{\{t\}}}_\text{$n_{0}\times m_{t}$} \\
	\underbrace{\mathbf{Z}^{[l]\{t\}}}_\text{$n_{l}\times m_{t}$} &= \underbrace{\underbrace{W^{[l]}}_\text{$n_{l}\times n_{l-1}$} \cdot \underbrace{\mathbf{A}^{[l-1]\{t\}}}_\text{$n_{l-1}\times m_{t}$}}_\text{$n_{l}\times m_{t}$} + \underbrace{\underbrace{b^{[l]}}_\text{$n_{l}\times 1$} \cdot \underbrace{\mathbf{1}^{T}}_\text{$1\times m_{t}$}}_\text{$n_{l}\times m_{t}$} \\
	\underbrace{\mathbf{A}^{[l]\{t\}}}_\text{$n_{l}\times m_{t}$} &= \sigma\left(\underbrace{\mathbf{Z}^{[l]\{t\}}}_\text{$n_{l}\times m_{t}$}\right),\;\;\;\;\;
	\underbrace{\hat{Y}^{\{t\}}}_\text{$n_{L}\times m_{t}$} = \underbrace{\mathbf{A}^{[L]\{t\}}}_\text{$n_{L}\times m_{t}$}
	\end{split}
	\label{forward_propagation_minibatch}
	\end{equation}
	where $\mathbf{1}$ is now an $m_{t}\times 1$ vector of ones. The errors in equation(\ref{delta_batch}) become:
	\begin{equation}
	\begin{split}
	\underbrace{\boldsymbol{\Delta}^{[L]\{t\}}}_\text{$n_{L}\times m_{t}$} &= \underbrace{-\left(Y^{\{t\}} - \hat{Y}^{\{t\}}\right)}_\text{$n_{L}\times m_{t}$} \odot \underbrace{\sigma'\left(\mathbf{Z}^{[L]\{t\}}\right)}_\text{$n_{L}\times m_{t}$} \\
	\underbrace{\boldsymbol{\Delta}^{[l]\{t\}}}_\text{$n_{l}\times m_{t}$} &= \underbrace{\left(\underbrace{W^{[l+1]T}}_\text{$n_{l}\times n_{l+1}$} \cdot \underbrace{\boldsymbol{\Delta}^{[l+1]\{t\}}}_\text{$n_{l+1}\times m_{t}$}\right)}_\text{$n_{l}\times m_{t}$} \odot \underbrace{\sigma'\left(\mathbf{Z}^{[l]\{t\}}\right)}_\text{$n_{l}\times m_{t}$}
	\end{split}
	\label{delta_minibatch}
	\end{equation}
	and the gradients in equation(\ref{dJ_dW_batch}) become:
	\begin{equation}
	\underbrace{\frac{\partial J^{\{t\}}}{\partial W^{[l]}}}_\text{$n_{l}\times n_{l-1}$} = \frac{1}{m_{t}}\cdot\underbrace{\underbrace{\boldsymbol{\Delta}^{[l]\{t\}}}_\text{$n_{l}\times m_{t}$} \cdot \underbrace{\mathbf{A}^{[l-1]\{t\}T}}_\text{$m_{t}\times n_{l-1}$}}_\text{$n_{l}\times n_{l-1}$},\;\;\;\;\;
	\underbrace{\frac{\partial J^{\{t\}}}{\partial b^{[l]}}}_\text{$n_{l}\times 1$} = \frac{1}{m_{t}}\cdot\underbrace{\underbrace{\boldsymbol{\Delta}^{[l]\{t\}}}_\text{$n_{l}\times m_{t}$} \cdot \underbrace{\mathbf{1}}_\text{$m_{t}\times 1$}}_\text{$n_{l}\times 1$}
	\label{dJ_dW_minibatch}
	\end{equation}
	\textbf{Note}: the weights $W^{[l]}$ and the biases $b^{[l]}$ do not have the superscript $\{t\}$, since all the mini-batches use the same weights and biases. Only the data and the values computed from the data change from one mini-batch to another. Also, the dimensions of the gradients are still $n_{l}\times n_{l-1}$ and $n_{l}\times 1$, i.e., they do not depend on the mini-batch size $m_{t}$.

	\subsection{Why Does It Work?}
	The mini-batches do not overlap, and together they contain all the $m$ examples. Hence, we can split the sum in the cost function of equation(\ref{cost_function}) into one sum for every mini-batch:
	\[
	J = \frac{1}{m}\sum_{i=1}^{m} E^{(i)} = \frac{1}{m}\sum_{t=1}^{N}\;\sum_{i\,\in\,\{t\}} E^{(i)}
	\]
	and from equation(\ref{minibatch_cost}), we know that $\sum_{i\,\in\,\{t\}} E^{(i)} = m_{t}\cdot J^{\{t\}}$, so:
	\begin{equation}
	J = \sum_{t=1}^{N} \frac{m_{t}}{m}\cdot J^{\{t\}}
	\label{cost_as_minibatch_average}
	\end{equation}
	i.e., the cost of the whole dataset is a weighted average of the costs of the mini-batches, where each mini-batch is weighted by its size. Since the derivative of a sum is the sum of the derivatives, the same is true for the gradients:
	\[
	\frac{\partial J}{\partial W^{[l]}} = \sum_{t=1}^{N} \frac{m_{t}}{m}\cdot\frac{\partial J^{\{t\}}}{\partial W^{[l]}}
	\]
	Therefore, the gradient of a single mini-batch is an \textbf{estimate} of the gradient of the whole dataset. Since the mini-batches are random samples, the gradient of a mini-batch is equal to the full gradient \textbf{on average}. The larger the mini-batch size $B$ is, the better the estimate is, but the more computation each update needs.
	\\\\
	\textbf{Note}: equation(\ref{cost_as_minibatch_average}) is exact only if all the mini-batches are evaluated with the \textbf{same} weights. In mini-batch gradient descent, the weights change after every mini-batch, so each $J^{\{t\}}$ is evaluated with different weights. This is another reason why the update of each mini-batch is only an estimate of the update of batch gradient descent.
	\\\\
	The mini-batch size $B$ controls which algorithm we get:
	\begin{center}
	\small
	\begin{tabular}{|c|c|>{\raggedright\arraybackslash}p{5cm}|}
		\hline
		\textbf{Mini-batch size} & \textbf{Updates per epoch} & \textbf{Algorithm} \\
		\hline
		$B = m$ & $N = 1$ & Batch gradient descent (Algorithm(\ref{backprop_algorithm})) \\
		$1 < B < m$ & $N = \lceil m/B \rceil$ & Mini-batch gradient descent (Algorithm(\ref{minibatch_algorithm})) \\
		$B = 1$ & $N = m$ & Stochastic gradient descent (SGD) \\
		\hline
	\end{tabular}
	\end{center}
	Common values of $B$ are $32$, $64$, $128$, and $256$. The code in Section(\ref{code_section}) uses $B = 128$.

	\subsection{The Mini-Batch Gradient Descent Algorithm}
	Algorithm(\ref{minibatch_algorithm}) shows the full mini-batch gradient descent. The lines inside the inner loop over the mini-batches are the same lines of Algorithm(\ref{backprop_algorithm}), applied to $X^{\{t\}}$ and $Y^{\{t\}}$ instead of $X$ and $Y$.

	\begin{algorithm}[htbp]
		\caption{Mini-Batch Gradient Descent with Backpropagation}
		\label{minibatch_algorithm}
		\begin{algorithmic}[1]
			\State $W^{[l]}$ = a randomly initialized matrix of size $n_{l} \times n_{l-1}$, for $l = 1, \dots, L$
			\State $b^{[l]}$ = a vector of zeros of size $n_{l} \times 1$, for $l = 1, \dots, L$
			\State $\alpha$ = the learning rate
			\State $B$ = the mini-batch size \label{alg2:batch_size}
			\State $T$ = the number of epochs
			\For {$e = 1$ to $T$}
				\State \textit{// shuffle and split} \label{alg2:shuffle_start}
				\State shuffle the columns of $X$ and $Y$ with the same random order \label{alg2:shuffle}
				\State split $X$ and $Y$ into $N = \lceil m/B \rceil$ mini-batches $X^{\{t\}}$ and $Y^{\{t\}}$ \label{alg2:split}
				\For {$t = 1$ to $N$} \label{alg2:minibatch_loop}
					\State $m_{t}$ = the number of columns of $X^{\{t\}}$ \label{alg2:m_t}
					\State \textit{// forward propagation} \label{alg2:forward_start}
					\State $\mathbf{A}^{[0]\{t\}} = X^{\{t\}}$ \label{alg2:A0}
					\For {$l = 1$ to $L$}
						\State $\mathbf{Z}^{[l]\{t\}} = W^{[l]} \cdot \mathbf{A}^{[l-1]\{t\}} + b^{[l]} \cdot \mathbf{1}^{T}$
						\State $\mathbf{A}^{[l]\{t\}} = \sigma\left(\mathbf{Z}^{[l]\{t\}}\right)$
					\EndFor
					\State $\hat{Y}^{\{t\}} = \mathbf{A}^{[L]\{t\}}$
					\State \textit{// backward propagation}
					\State $\boldsymbol{\Delta}^{[L]\{t\}} = -\left(Y^{\{t\}} - \hat{Y}^{\{t\}}\right) \odot \sigma'\left(\mathbf{Z}^{[L]\{t\}}\right)$ \label{alg2:delta_L}
					\For {$l = L$ down to $1$}
						\State $\nabla_{W^{[l]}} = \frac{1}{m_{t}}\cdot\boldsymbol{\Delta}^{[l]\{t\}} \cdot \mathbf{A}^{[l-1]\{t\}T}$ \label{alg2:grad_W}
						\State $\nabla_{b^{[l]}} = \frac{1}{m_{t}}\cdot\boldsymbol{\Delta}^{[l]\{t\}} \cdot \mathbf{1}$ \label{alg2:grad_b}
						\If {$l > 1$}
							\State $\boldsymbol{\Delta}^{[l-1]\{t\}} = \left(W^{[l]T} \cdot \boldsymbol{\Delta}^{[l]\{t\}}\right) \odot \sigma'\left(\mathbf{Z}^{[l-1]\{t\}}\right)$
						\EndIf
					\EndFor
					\State \textit{// update} \label{alg2:update_start}
					\For {$l = 1$ to $L$}
						\State $W^{[l]} \leftarrow W^{[l]} - \alpha \cdot \nabla_{W^{[l]}}$
						\State $b^{[l]} \leftarrow b^{[l]} - \alpha \cdot \nabla_{b^{[l]}}$
					\EndFor \label{alg2:update_end}
				\EndFor \label{alg2:minibatch_loop_end}
			\EndFor
		\end{algorithmic}
	\end{algorithm}

	\subsection{Changes Compared to Batch Gradient Descent}
	\label{minibatch_changes}
	If we compare Algorithm(\ref{minibatch_algorithm}) with Algorithm(\ref{backprop_algorithm}), we can see that the equations of the forward propagation, the backward propagation, and the update rule are exactly the same. Only the following things change:
	\begin{enumerate}[leftmargin=1cm, label=\arabic*.]
		\item \textbf{A new parameter}: the mini-batch size $B$ in line~\ref{alg2:batch_size} of Algorithm(\ref{minibatch_algorithm}).
		\item \textbf{Shuffling and splitting} (new lines~\ref{alg2:shuffle_start}--\ref{alg2:split}): at the beginning of every epoch, we shuffle the columns of $X$ and $Y$ and split them into $N$ mini-batches as in equation(\ref{number_of_minibatches}). Algorithm(\ref{backprop_algorithm}) does not need these lines.
		\item \textbf{A new inner loop} (lines~\ref{alg2:minibatch_loop}--\ref{alg2:minibatch_loop_end}): the whole body of the epoch loop of Algorithm(\ref{backprop_algorithm}), i.e., lines~\ref{alg1:forward_start}--\ref{alg1:update_end}, is moved inside a loop over the $N$ mini-batches.
		\item \textbf{The data of a mini-batch}: line~\ref{alg2:A0} uses $X^{\{t\}}$ instead of $X$ in line~\ref{alg1:A0} of Algorithm(\ref{backprop_algorithm}), and line~\ref{alg2:delta_L} uses $Y^{\{t\}}$ instead of $Y$ in line~\ref{alg1:delta_L}. Hence, all the matrices $\mathbf{Z}^{[l]\{t\}}$, $\mathbf{A}^{[l]\{t\}}$, and $\boldsymbol{\Delta}^{[l]\{t\}}$ have $m_{t}$ columns instead of $m$ columns.
		\item \textbf{The average over a mini-batch}: lines~\ref{alg2:grad_W} and~\ref{alg2:grad_b} divide by the size of the mini-batch $m_{t}$ (computed in line~\ref{alg2:m_t}) instead of dividing by $m$ in lines~\ref{alg1:grad_W} and~\ref{alg1:grad_b}. We use $m_{t}$ instead of $B$ because the last mini-batch may be smaller than $B$.
		\item \textbf{The number of updates}: the update (lines~\ref{alg2:update_start}--\ref{alg2:update_end}) is inside the loop over the mini-batches, so the weights are updated $N$ times in every epoch instead of only one time.
	\end{enumerate}
	\textbf{Note}: if we set $B = m$ in Algorithm(\ref{minibatch_algorithm}), we will get $N = 1$ and $m_{1} = m$, i.e., a single mini-batch that contains the whole dataset. In this case, Algorithm(\ref{minibatch_algorithm}) is exactly the same as Algorithm(\ref{backprop_algorithm}), except for the shuffling, which has no effect since the sum over all the examples does not depend on their order.

	\newpage
	\section{Implementation in NumPy}
	\label{code_section}
	In this section, we will link every equation of this paper to its line in a NumPy implementation of the network~\cite{alasmary_nn_numpy}. We use the notebook \nolinkurl{simplest_neural_network.ipynb}, which contains the minimal code of a network with $3$ layers. For every part, we write the equation again with its number, and then we show the code that computes it. The line numbers of the code are the line numbers inside the cells of the notebook, so that the reader can find every line easily.
	\\\\
	The network is trained on the AHDBase dataset of handwritten Arabic digits, which contains $60,000$ training images and $10,000$ test images. Each image has $8\times 8 = 64$ pixels, and each label is one of the $10$ digits written as a \textbf{one-hot} vector, i.e., a vector of $10$ zeros with a single $1$ in the position of the digit. Hence, the network has the following sizes:
	\[
	L = 3,\;\;\;\;\; n_{0} = 64,\;\;\;\;\; n_{1} = 32,\;\;\;\;\; n_{2} = 16,\;\;\;\;\; n_{3} = 10,\;\;\;\;\; m = 60000
	\]
	which is the same as our running example in Figure(\ref{network_figure}), but with more nodes in each layer.

	\subsection{The Variables}
	The following table shows the name of every variable of the paper in the code:
	\begin{center}
	\begin{tabular}{|c|c|c|}
		\hline
		\textbf{Paper} & \textbf{Code} & \textbf{Dimensions in the code} \\
		\hline
		$X$, $Y$ & \texttt{X\_train}, \texttt{y\_train} & $64\times 60000$, $10\times 60000$ \\
		$\mathbf{A}^{[0]}$ & \texttt{A0} & $64\times m$ \\
		$W^{[1]}$, $W^{[2]}$, $W^{[3]}$ & \texttt{W1}, \texttt{W2}, \texttt{W3} & $32\times 64$, $16\times 32$, $10\times 16$ \\
		$b^{[1]}$, $b^{[2]}$, $b^{[3]}$ & \texttt{b1}, \texttt{b2}, \texttt{b3} & $32\times 1$, $16\times 1$, $10\times 1$ \\
		$\mathbf{Z}^{[l]}$, $\mathbf{A}^{[l]}$ & \texttt{Z1}, \texttt{A1}, \dots, \texttt{Z3}, \texttt{A3} & $n_{l}\times m$ \\
		$\hat{Y}$, $Y$ & \texttt{y\_pred}, \texttt{y\_true} & $10\times m$ \\
		$\boldsymbol{\Delta}^{[l]}$ & \texttt{delta1}, \texttt{delta2}, \texttt{delta3} & $n_{l}\times m$ \\
		$\frac{\partial J}{\partial W^{[l]}}$, $\frac{\partial J}{\partial b^{[l]}}$ & \texttt{dW1}, \dots, \texttt{db3} & same as \texttt{W1}, \dots, \texttt{b3} \\
		$J$ & \texttt{cost} & $1\times 1$ \\
		$\alpha$ & \texttt{alpha} & $1\times 1$ \\
		$m$ (or $m_{t}$) & \texttt{m} & $1\times 1$ \\
		$B$ & \texttt{batch\_size} & $1\times 1$ \\
		\hline
	\end{tabular}
	\end{center}
	and the following table shows the operations:
	\begin{center}
	\begin{tabular}{|c|c|c|}
		\hline
		\textbf{Operation} & \textbf{Paper} & \textbf{Code} \\
		\hline
		Matrix product & $W^{[1]} \cdot \mathbf{A}^{[0]}$ & \texttt{W1.dot(A0)} \\
		Hadamard product & $U \odot V$ & \texttt{U * V} \\
		Transpose & $W^{[3]T}$ & \texttt{W3.T} \\
		Sum of the columns & $\boldsymbol{\Delta}^{[l]} \cdot \mathbf{1}$ & \texttt{delta1.sum(axis=1, keepdims=True)} \\
		\hline
	\end{tabular}
	\end{center}

	\subsection{The Examples as Columns}
	In Section(\ref{batch_section}), we stacked the examples as \textbf{columns}, so that $X$ has the dimensions $n_{0}\times m$:
	\[
	\underbrace{X}_\text{$n_{0}\times m$} =
	\underbrace{
		\begin{bmatrix}
		|       & |       &        & |       \\
		x^{(1)} & x^{(2)} & \dots  & x^{(m)} \\
		|       & |       &        & |
		\end{bmatrix}}_\text{$n_{0}\times m$}
	\]
	The dataset is loaded from a CSV file in which every example is a \textbf{row}, so the notebook takes the transpose of the data before training:
	\begin{code}
# Take transpose since the neural network considers columns as examples and rows as features
# i.e. if m is the number of samples (images) and n is the number of features (image pixels), then
# n x m is the shape that is accepted by the network
X_train = X_train.T
y_train = y_train.T
\end{code}
	After this step, \texttt{X\_train} has the dimensions $64\times 60000$ and \texttt{y\_train} has the dimensions $10\times 60000$.

	\subsection{The Sigmoid Function and Its Derivative}
	The sigmoid function of equation(\ref{sigmoid}):
	\[
	\sigma(z) = \frac{1}{1 + e^{-z}}
	\]
	and its derivative of equation(\ref{sigmoid_derivative}):
	\[
	\sigma'(z) = \sigma(z)\cdot\left(1 - \sigma(z)\right)
	\]
	are implemented as follows:
	\begin{code}
def sigmoid(Z):
    return 1. / (1 + np.exp(-Z))

def sigmoid_prime(A):
    return A * (1 - A)
\end{code}
	\textbf{Note}: \texttt{sigmoid\_prime} takes the \textbf{output} $A$ of the layer, not the weighted sum $Z$. Since $\mathbf{A}^{[l]} = \sigma\left(\mathbf{Z}^{[l]}\right)$ from equation(\ref{forward_propagation_batch}), we can write equation(\ref{sigmoid_derivative}) as:
	\[
	\sigma'\left(\mathbf{Z}^{[l]}\right) = \sigma\left(\mathbf{Z}^{[l]}\right) \odot \left(1 - \sigma\left(\mathbf{Z}^{[l]}\right)\right) = \mathbf{A}^{[l]} \odot \left(1 - \mathbf{A}^{[l]}\right)
	\]
	so we can reuse the outputs that we already computed in the forward propagation instead of computing the exponentials again. Hence, \texttt{sigmoid\_prime(A2)} is $\sigma'\left(\mathbf{Z}^{[2]}\right)$, and \texttt{sigmoid\_prime(y\_pred)} is $\sigma'\left(\mathbf{Z}^{[3]}\right)$ since $\hat{Y} = \mathbf{A}^{[3]}$. Both functions work element-wise on matrices, as we explained in Section(\ref{sigmoid_section}).

	\subsection{Initializing The Weights}
	Lines~\ref{alg1:init_W} and~\ref{alg1:init_b} of Algorithm(\ref{backprop_algorithm}) initialize the weights and the biases. The weights $W^{[l]}$ have the dimensions $n_{l}\times n_{l-1}$ as we saw in Section(\ref{forward_section}):
	\[
	\underbrace{W^{[1]}}_\text{$32\times 64$},\;\;\;\;\;
	\underbrace{W^{[2]}}_\text{$16\times 32$},\;\;\;\;\;
	\underbrace{W^{[3]}}_\text{$10\times 16$},\;\;\;\;\;
	\underbrace{b^{[1]}}_\text{$32\times 1$},\;\;\;\;\;
	\underbrace{b^{[2]}}_\text{$16\times 1$},\;\;\;\;\;
	\underbrace{b^{[3]}}_\text{$10\times 1$}
	\]
	\begin{code}
W1 = 2*np.random.random((32, 64)) - 1
W1 /= 32

b1 = 2*np.random.random((32, 1)) - 1

W2 = 2*np.random.random((16, 32)) - 1
W2 /= 16

b2 = 2*np.random.random((16, 1)) - 1

W3 = 2*np.random.random((10, 16)) - 1
W3 /= 10

b3 = 2*np.random.random((10, 1)) - 1
\end{code}
	The function \texttt{np.random.random} gives random numbers between $0$ and $1$. Multiplying them by $2$ and subtracting $1$ gives random numbers between $-1$ and $1$. The weights are then divided by $n_{l}$ to make them small. Small weights keep the weighted sums $\mathbf{Z}^{[l]}$ close to zero, where the sigmoid function is not flat and its derivative $\sigma'$ is not close to zero. If $\sigma'$ were close to zero, the errors $\boldsymbol{\Delta}^{[l]}$ in equation(\ref{delta_batch}) would also be close to zero and the network would learn very slowly.
	\\\\
	\textbf{Note}: the biases are stored as column-vectors of shape $(n_{l}, 1)$, as we explained after equation(\ref{forward_propagation_batch}). Also, the code initializes the biases randomly, while Algorithm(\ref{backprop_algorithm}) initializes them with zeros. Both choices work, since the random weights are already enough to make the nodes of each layer different from each other.

	\subsection{Forward Propagation}
	The forward propagation of equation(\ref{forward_propagation_batch}):
	\[
	\begin{split}
	\underbrace{\mathbf{A}^{[0]}}_\text{$n_{0}\times m$} &= \underbrace{X}_\text{$n_{0}\times m$} \\
	\underbrace{\mathbf{Z}^{[l]}}_\text{$n_{l}\times m$} &= \underbrace{W^{[l]}}_\text{$n_{l}\times n_{l-1}$} \cdot \underbrace{\mathbf{A}^{[l-1]}}_\text{$n_{l-1}\times m$} + \underbrace{b^{[l]}}_\text{$n_{l}\times 1$} \cdot \underbrace{\mathbf{1}^{T}}_\text{$1\times m$},\;\;\;\;\;
	\underbrace{\mathbf{A}^{[l]}}_\text{$n_{l}\times m$} = \sigma\left(\underbrace{\mathbf{Z}^{[l]}}_\text{$n_{l}\times m$}\right)
	\end{split}
	\]
	and $\hat{Y} = \mathbf{A}^{[3]}$ (lines~\ref{alg1:A0}--\ref{alg1:Yhat} of Algorithm(\ref{backprop_algorithm})) are computed in lines $16$--$29$ of the batch gradient descent cell:
	\begin{code}[firstnumber=16]
    A0 = X_train
    y_true = y_train

    # feedforward
    Z1 = W1.dot(A0) + b1
    A1 = sigmoid(Z1)

    Z2 = W2.dot(A1) + b2
    A2 = sigmoid(Z2)

    Z3 = W3.dot(A2) + b3
    A3 = sigmoid(Z3)

    y_pred = A3
\end{code}
	The loop over the layers in Algorithm(\ref{backprop_algorithm}) is written as three separate steps, since the network has only $3$ layers. The term $b^{[l]}\cdot\mathbf{1}^{T}$ is written as \texttt{+ b1}, since NumPy \textbf{broadcasts} the bias vector to all the $m$ columns, as we explained after equation(\ref{forward_propagation_batch}).

	\subsection{The Cost}
	The cost function of equation(\ref{cost_function}) is:
	\[
	\begin{split}
	J &= \frac{1}{m}\sum_{i=1}^{m} E^{(i)} = \frac{1}{m}\sum_{i=1}^{m} \frac{1}{2}\cdot\left(y^{(i)} - \hat{y}^{(i)}\right)^{T}\cdot\left(y^{(i)} - \hat{y}^{(i)}\right) \\
	&= \frac{1}{2m}\sum_{i=1}^{m}\sum_{j=1}^{n_{L}}\left(y^{(i)}_{j} - \hat{y}^{(i)}_{j}\right)^{2}
	\end{split}
	\]
	The double sum goes over all the elements of the matrix $\left(Y - \hat{Y}\right)$: the sum over $j$ goes over its rows (the output nodes) and the sum over $i$ goes over its columns (the examples). \texttt{np.sum} adds up all the elements of a matrix, so it computes both sums at once in line $32$:
	\begin{code}[firstnumber=31]
    y_true = np.atleast_2d(y_true)
    cost = 0.5 * np.sum((y_true - y_pred) ** 2) / m
\end{code}
	where \texttt{** 2} squares every element of the matrix. Line $31$ only makes sure that \texttt{y\_true} is a 2-D array. The cost is not needed to compute the gradients, but it is printed and saved to see whether the network is learning.

	\subsection{Backward Propagation}
	The errors of equation(\ref{delta_batch}):
	\[
	\begin{split}
	\underbrace{\boldsymbol{\Delta}^{[L]}}_\text{$n_{L}\times m$} &= \underbrace{-\left(Y - \hat{Y}\right)}_\text{$n_{L}\times m$} \odot \underbrace{\sigma'\left(\mathbf{Z}^{[L]}\right)}_\text{$n_{L}\times m$} \\
	\underbrace{\boldsymbol{\Delta}^{[l]}}_\text{$n_{l}\times m$} &= \underbrace{\left(\underbrace{W^{[l+1]T}}_\text{$n_{l}\times n_{l+1}$} \cdot \underbrace{\boldsymbol{\Delta}^{[l+1]}}_\text{$n_{l+1}\times m$}\right)}_\text{$n_{l}\times m$} \odot \underbrace{\sigma'\left(\mathbf{Z}^{[l]}\right)}_\text{$n_{l}\times m$}
	\end{split}
	\]
	(lines~\ref{alg1:delta_L} and~\ref{alg1:delta_l} of Algorithm(\ref{backprop_algorithm})) are computed in lines $35$--$37$:
	\begin{code}[firstnumber=34]
    # backpropagate error
    delta3 = -(y_true - y_pred) * sigmoid_prime(y_pred)
    delta2 = W3.T.dot(delta3) * sigmoid_prime(A2)
    delta1 = W2.T.dot(delta2) * sigmoid_prime(A1)
\end{code}
	Line $35$ is the error of the last layer with $L = 3$, i.e., the \textbf{loss derivative} $-\left(Y - \hat{Y}\right)$ multiplied element-wise by the \textbf{activation's derivative} $\sigma'\left(\mathbf{Z}^{[3]}\right)$ as in equation(\ref{general_delta_L}). Lines $36$ and $37$ are the errors of the hidden layers with $l = 2$ and $l = 1$: the \textbf{weighted propagated errors from the next layer} $W^{[l+1]T}\cdot\boldsymbol{\Delta}^{[l+1]}$ multiplied element-wise by the activation's derivative as in equation(\ref{general_delta_l}). For example, line $36$ is:
	\[
	\underbrace{\boldsymbol{\Delta}^{[2]}}_\text{$16\times m$} = \underbrace{\left(\underbrace{W^{[3]T}}_\text{$16\times 10$} \cdot \underbrace{\boldsymbol{\Delta}^{[3]}}_\text{$10\times m$}\right)}_\text{$16\times m$} \odot \underbrace{\sigma'\left(\mathbf{Z}^{[2]}\right)}_\text{$16\times m$}
	\]
	The errors are computed from the last layer to the first layer, since each line needs the error of the next layer. This is the \textbf{back}propagation.

	\subsection{The Gradients}
	The gradients of equation(\ref{dJ_dW_batch}):
	\[
	\underbrace{\frac{\partial J}{\partial W^{[l]}}}_\text{$n_{l}\times n_{l-1}$} = \frac{1}{m}\cdot\underbrace{\underbrace{\boldsymbol{\Delta}^{[l]}}_\text{$n_{l}\times m$} \cdot \underbrace{\mathbf{A}^{[l-1]T}}_\text{$m\times n_{l-1}$}}_\text{$n_l\times n_{l-1}$},\;\;\;\;\;
	\underbrace{\frac{\partial J}{\partial b^{[l]}}}_\text{$n_{l}\times 1$} = \frac{1}{m}\cdot\underbrace{\underbrace{\boldsymbol{\Delta}^{[l]}}_\text{$n_{l}\times m$} \cdot \underbrace{\mathbf{1}}_\text{$m\times 1$}}_\text{$n_l\times 1$}
	\]
	(lines~\ref{alg1:grad_W} and~\ref{alg1:grad_b} of Algorithm(\ref{backprop_algorithm})) are computed in lines $39$--$45$:
	\begin{code}[firstnumber=39]
    dW1 = delta1.dot(A0.T) / m
    dW2 = delta2.dot(A1.T) / m
    dW3 = delta3.dot(A2.T) / m

    db1 = delta1.sum(axis=1, keepdims=True) / m
    db2 = delta2.sum(axis=1, keepdims=True) / m
    db3 = delta3.sum(axis=1, keepdims=True) / m
\end{code}
	Lines $39$--$41$ are the \textbf{sum of outer products} that we wrote as a matrix multiplication in Section(\ref{batch_section}): the error of each layer multiplied by the transposed input of the same layer. For example, line $39$ is:
	\[
	\underbrace{\frac{\partial J}{\partial W^{[1]}}}_\text{$32\times 64$} = \frac{1}{m}\cdot\underbrace{\underbrace{\boldsymbol{\Delta}^{[1]}}_\text{$32\times m$} \cdot \underbrace{\mathbf{A}^{[0]T}}_\text{$m\times 64$}}_\text{$32\times 64$}
	\]
	Lines $43$--$45$ add up the $m$ columns of each error with \texttt{sum(axis=1)}, which is the same as $\boldsymbol{\Delta}^{[l]}\cdot\mathbf{1}$. The option \texttt{keepdims=True} keeps the result as a column-vector of shape $(n_{l}, 1)$ instead of a 1-D array of shape $(n_{l},)$, so that it has the same shape as the bias.
	\\\\
	\textbf{Note}: Algorithm(\ref{backprop_algorithm}) computes the gradients of each layer right after its error, while the code computes all the errors first and then all the gradients. Both orders give the same result, since the gradients are not used to compute the errors.

	\subsection{Checking The Dimensions}
	Throughout this paper, we showed the dimensions under each term. In particular, the gradients have the same dimensions as the weights and the biases, as in equations(\ref{general_dE_dW}) and (\ref{general_dE_db}):
	\[
	\underbrace{\frac{\partial J}{\partial W^{[l]}}}_\text{$n_{l}\times n_{l-1}$}\;\;\text{and}\;\;\underbrace{W^{[l]}}_\text{$n_{l}\times n_{l-1}$},\;\;\;\;\;
	\underbrace{\frac{\partial J}{\partial b^{[l]}}}_\text{$n_{l}\times 1$}\;\;\text{and}\;\;\underbrace{b^{[l]}}_\text{$n_{l}\times 1$}
	\]
	The code checks the same thing in lines $48$--$53$. If any shape is different, the program stops with the message \texttt{'Size Mismatch'}:
	\begin{code}[firstnumber=48]
    assert W1.shape == dW1.shape, 'Size Mismatch'
    assert W2.shape == dW2.shape, 'Size Mismatch'
    assert W3.shape == dW3.shape, 'Size Mismatch'
    assert b1.shape == db1.shape, 'Size Mismatch'
    assert b2.shape == db2.shape, 'Size Mismatch'
    assert b3.shape == db3.shape, 'Size Mismatch'
\end{code}
	This is the same idea as the underbraces of this paper: checking the dimensions is an easy way to find mistakes in the equations and in the code.

	\subsection{Updating The Weights}
	The update rule of equation(\ref{update_rule}):
	\[
	\begin{split}
	W^{[l]}_{new} &= W^{[l]}_{old} - \alpha\cdot\frac{\partial J}{\partial W^{[l]}} \\
	b^{[l]}_{new} &= b^{[l]}_{old} - \alpha\cdot\frac{\partial J}{\partial b^{[l]}}
	\end{split}
	\]
	(lines~\ref{alg1:update_W} and~\ref{alg1:update_b} of Algorithm(\ref{backprop_algorithm})) is applied in lines $56$--$62$:
	\begin{code}[firstnumber=55]
    # update weights
    W1 = W1 - alpha*dW1
    W2 = W2 - alpha*dW2
    W3 = W3 - alpha*dW3

    b1 = b1 - alpha*db1
    b2 = b2 - alpha*db2
    b3 = b3 - alpha*db3
\end{code}
	All the gradients are computed before any weight is updated, so all the errors are computed with the \textbf{old} weights, as we noted after equation(\ref{update_rule}).

	\subsection{The Training Loop (Batch Gradient Descent)}
	Finally, the epoch loop of line~\ref{alg1:epoch_loop} of Algorithm(\ref{backprop_algorithm}) is written in lines $1$--$14$ of the cell:
	\begin{code}
m = X_train.shape[1]
alpha = 0.9
costs = []
tic = time.time()
for epoch in range(20000):

    ### Shuffling ###
    indices = np.arange(X_train.shape[1])
    np.random.shuffle(indices)

    X_train = X_train[:, indices]
    y_train = y_train[:, indices]

    #################
\end{code}
	Line $1$ gets the number of examples $m$, which is the number of columns of $X$. Line $2$ sets the learning rate $\alpha$, and line $5$ sets the number of epochs $T = 20000$. All the lines that we showed above (lines $16$--$62$) are inside this loop, so the weights are updated once in every epoch.
	\\\\
	Lines $8$--$12$ shuffle the columns of $X$ and $Y$ with the same random order \texttt{indices}. As we noted at the end of Section(\ref{minibatch_changes}), shuffling has no effect in batch gradient descent, since the gradient adds up all the examples and the sum does not depend on their order. It becomes important in the mini-batch version below.

	\subsection{The Mini-Batch Version}
	The notebook also contains the mini-batch gradient descent of Algorithm(\ref{minibatch_algorithm}). The splitting into mini-batches of line~\ref{alg2:split} of Algorithm(\ref{minibatch_algorithm}) and equation(\ref{number_of_minibatches}) is done by the following function:
	\begin{code}
def get_mini_batches(X, y, batch_size):
    assert X.shape[1] == y.shape[1], 'Size Mismatch'
    m = X.shape[1]
    num_of_mini_batches = m // batch_size

    for i in range(num_of_mini_batches):
        start = i * batch_size
        end = start + batch_size
        mini_bacth_X = X[:, start:end]
        mini_batch_y = y[:, start:end]
        yield mini_bacth_X, mini_batch_y

    if m % batch_size != 0:
        last_mini_bacth_X = X[:, end:]
        last_mini_bacth_y = y[:, end:]
        yield last_mini_bacth_X, last_mini_bacth_y
\end{code}
	Line $4$ computes the number of \textbf{full} mini-batches $\lfloor m/B \rfloor$, where \texttt{//} is the integer division that rounds down. Lines $6$--$11$ return the columns from \texttt{start} to \texttt{end} of each full mini-batch, i.e., $X^{\{t\}}$ and $Y^{\{t\}}$ with $m_{t} = B$ columns. If $m$ is not divisible by $B$, lines $13$--$16$ return the remaining columns as the last, smaller mini-batch with $m_{N} = m - (N-1)\cdot B$ columns. Hence, the total number of mini-batches is $N = \lceil m/B \rceil$, as in equation(\ref{number_of_minibatches}). For our dataset, $N = \lceil 60000/128 \rceil = 469$, where the first $468$ mini-batches have $128$ examples and the last one has $60000 - 468\cdot 128 = 96$ examples.
	\\\\
	The training loop of the mini-batch version is shown below. We only show the lines that differ from the batch version, since the forward propagation, the backward propagation, the gradients, and the update are the same as above with one more level of indentation:
	\begin{code}[firstnumber=1]
alpha = 0.9
costs = []
batch_size = 128

tic = time.time()
for epoch in range(1000):
    total_cost = 0

    ### Shuffling ###
    indices = np.arange(X_train.shape[1])
    np.random.shuffle(indices)

    X_train = X_train[:, indices]
    y_train = y_train[:, indices]

    #################
    j = 0
    for mini_bacth_X, mini_bacth_y in get_mini_batches(X_train, y_train, batch_size):
        m = mini_bacth_y.shape[1]
        y_true = mini_bacth_y
        A0 = mini_bacth_X
\end{code}
	\begin{code}[firstnumber=39]
        # backpropagate error
        delta3 = (y_pred - y_true) * sigmoid_prime(y_pred)
\end{code}
	These lines follow the changes of Section(\ref{minibatch_changes}):
	\begin{enumerate}[leftmargin=1cm, label=\arabic*.]
		\item Line $3$ sets the mini-batch size $B = 128$ (line~\ref{alg2:batch_size} of Algorithm(\ref{minibatch_algorithm})).
		\item Lines $10$--$14$ shuffle the columns of $X$ and $Y$ at the beginning of every epoch (line~\ref{alg2:shuffle}). Here, the shuffling is important, since it gives different mini-batches in every epoch.
		\item Line $18$ is the new inner loop over the mini-batches (line~\ref{alg2:minibatch_loop}), and \texttt{get\_mini\_batches} does the splitting (line~\ref{alg2:split}).
		\item Line $19$ gets the size of the current mini-batch $m_{t}$ (line~\ref{alg2:m_t}). The code keeps the variable name \texttt{m}, so all the lines \texttt{/ m} of the gradients now divide by $m_{t}$, as in equation(\ref{dJ_dW_minibatch}). This is also correct for the last, smaller mini-batch.
		\item Lines $20$ and $21$ use the mini-batch $Y^{\{t\}}$ and $X^{\{t\}}$ instead of the whole dataset (line~\ref{alg2:A0}).
		\item The update of the weights (lines $60$--$67$ of this cell) is inside the inner loop, so the weights are updated $469$ times in every epoch instead of only one time.
	\end{enumerate}
	\textbf{Note}: line $40$ writes the loss derivative as $\left(\hat{Y} - Y\right)$ instead of $-\left(Y - \hat{Y}\right)$. The two are the same, since $-\left(Y - \hat{Y}\right) = \hat{Y} - Y$.
	\\\\
	\textbf{Note}: the notebook also contains a stochastic gradient descent version, which is the special case $B = 1$. It loops over the columns of $X$ one by one and updates the weights after every example. Since $m_{t} = 1$, its gradients do not need to be divided by $m_{t}$.

	\subsection{The Complete Implementation}
	\label{complete_code_section}
	Finally, the following code block shows the whole implementation in one place. It collects the cells of the notebook~\cite{alasmary_nn_numpy} in the same order as in this section: loading and preparing the data (Section(\ref{code_section})), the helper functions, the initialization of the weights, the mini-batch gradient descent of Algorithm(\ref{minibatch_algorithm}), and the evaluation on the test set. We removed the cells that only display the data, and the notebook command \texttt{\%matplotlib inline}, so that the code can also run as a normal Python script. The files \texttt{AHDBase\_TrainingSet.csv} and \texttt{AHDBase\_TestingSet.csv} from the repository must be in the same folder as the script.
	\\\\
	\textbf{Note}: since the batch gradient descent and the stochastic gradient descent are special cases of the mini-batch gradient descent (Section(\ref{minibatch_section})), we can get Algorithm(\ref{backprop_algorithm}) by setting \texttt{batch\_size = X\_train.shape[1]}, i.e., $B = m$, and the stochastic gradient descent by setting \texttt{batch\_size = 1}, i.e., $B = 1$.
	\begin{lstlisting}
# ---------------- imports ----------------
from sklearn.model_selection import train_test_split
from sklearn.preprocessing import LabelBinarizer
from sklearn.metrics import classification_report
import numpy as np
import pandas as pd
import matplotlib.pyplot as plt
import time


# ---------------- load the data ----------------
data = pd.read_csv('AHDBase_TrainingSet.csv')
data_cols = list(data.columns)
data_cols.remove('Unnamed: 0')
train_data = data[data_cols]

data = pd.read_csv('AHDBase_TestingSet.csv')
data_cols = list(data.columns)
data_cols.remove('Unnamed: 0')
test_data = data[data_cols]

output = 'label'
features = list(train_data.columns)
features.remove(output)


# ---------------- prepare the data ----------------
# convert labels into one-hot encoding vectors
binarizer = LabelBinarizer()
one_hot_encoded_train_labels = binarizer.fit_transform(train_data[output])

X_train = train_data[features].values
y_train = one_hot_encoded_train_labels

X_test = test_data[features].values
y_test = binarizer.transform(test_data[output])

# Take transpose since the neural network considers columns as examples and rows as features
# i.e. if m is the number of samples (images) and n is the number of features (image pixels), then 
# n x m is the shape that is accepted by the network
X_train = X_train.T
y_train = y_train.T


# ---------------- helper functions ----------------
def sigmoid(Z):
    return 1. / (1 + np.exp(-Z))

def sigmoid_prime(A):
    return A * (1 - A)

def get_mini_batches(X, y, batch_size):
    assert X.shape[1] == y.shape[1], 'Size Mismatch'
    m = X.shape[1]
    num_of_mini_batches = m // batch_size

    for i in range(num_of_mini_batches):
        start = i * batch_size
        end = start + batch_size
        mini_bacth_X = X[:, start:end]
        mini_batch_y = y[:, start:end]
        yield mini_bacth_X, mini_batch_y

    if m % batch_size != 0:
        last_mini_bacth_X = X[:, end:]
        last_mini_bacth_y = y[:, end:]
        yield last_mini_bacth_X, last_mini_bacth_y


# ---------------- initialize the weights randomly ----------------
W1 = 2*np.random.random((32, 64)) - 1
W1 /= 32

b1 = 2*np.random.random((32, 1)) - 1

W2 = 2*np.random.random((16, 32)) - 1
W2 /= 16

b2 = 2*np.random.random((16, 1)) - 1

W3 = 2*np.random.random((10, 16)) - 1
W3 /= 10

b3 = 2*np.random.random((10, 1)) - 1

print(W1.shape, b1.shape)
print(W2.shape, b2.shape)
print(W3.shape, b3.shape)


# ---------------- mini-batch gradient descent ----------------
alpha = 0.9
costs = []
batch_size = 128

tic = time.time()
for epoch in range(1000):
    total_cost = 0
    
    ### Shuffling ###
    indices = np.arange(X_train.shape[1])
    np.random.shuffle(indices)

    X_train = X_train[:, indices]
    y_train = y_train[:, indices]
    
    #################
    j = 0
    for mini_bacth_X, mini_bacth_y in get_mini_batches(X_train, y_train, batch_size):
        m = mini_bacth_y.shape[1]
        y_true = mini_bacth_y
        A0 = mini_bacth_X
        
        # feedforward
        Z1 = W1.dot(A0) + b1
        A1 = sigmoid(Z1)

        Z2 = W2.dot(A1) + b2
        A2 = sigmoid(Z2)

        Z3 = W3.dot(A2) + b3
        A3 = sigmoid(Z3)

        y_pred = A3

        y_true = np.atleast_2d(y_true)
        cost = 0.5 * np.sum((y_true - y_pred) ** 2) / m
        total_cost += cost

        # backpropagate error
        delta3 = (y_pred - y_true) * sigmoid_prime(y_pred)
        delta2 = W3.T.dot(delta3) * sigmoid_prime(A2)
        delta1 = W2.T.dot(delta2) * sigmoid_prime(A1)

        dW1 = delta1.dot(A0.T) / m
        dW2 = delta2.dot(A1.T) / m
        dW3 = delta3.dot(A2.T) / m

        db1 = delta1.sum(axis=1, keepdims=True) / m
        db2 = delta2.sum(axis=1, keepdims=True) / m
        db3 = delta3.sum(axis=1, keepdims=True) / m


        assert W1.shape == dW1.shape, 'Size Mismatch'
        assert W2.shape == dW2.shape, 'Size Mismatch'
        assert W3.shape == dW3.shape, 'Size Mismatch'
        assert b1.shape == db1.shape, 'Size Mismatch'
        assert b2.shape == db2.shape, 'Size Mismatch'
        assert b3.shape == db3.shape, 'Size Mismatch'

        # update weights
        W1 = W1 - alpha*dW1
        W2 = W2 - alpha*dW2
        W3 = W3 - alpha*dW3

        b1 = b1 - alpha*db1
        b2 = b2 - alpha*db2
        b3 = b3 - alpha*db3
        
        j += 1
        costs.append(cost)
    
    if epoch % 10 == 0:
        print(epoch, total_cost / j)
        
toc = time.time()


# ---------------- evaluate on the test set ----------------
A0 = X_test.T
y_true = y_test.T

# feedforward
Z1 = W1.dot(A0) + b1
A1 = sigmoid(Z1)

Z2 = W2.dot(A1) + b2
A2 = sigmoid(Z2)

Z3 = W3.dot(A2) + b3
A3 = sigmoid(Z3)

y_pred = A3

my_labels_pred = binarizer.inverse_transform(y_pred.T)
print(classification_report(test_data[output], my_labels_pred))

plt.plot(costs)
\end{lstlisting}

\end{document}